\RequirePackage{booktabs}
\documentclass[sn-vancouver,Numbered]{sn-jnl}% Math and Physical Sciences Numbered Reference Style 
\usepackage{graphicx}%
\usepackage{multirow}%
\usepackage{amsmath,amssymb,amsfonts}%
\usepackage{amsthm}%
\usepackage{mathrsfs}%
\usepackage[title]{appendix}%
\usepackage{xcolor}%
\usepackage{textcomp}%
\usepackage{manyfoot}%
\usepackage{booktabs}%
\usepackage{algorithm}%
\usepackage{algorithmicx}%
\usepackage{algpseudocode}%
\usepackage{listings}%

\begin{document}
	
	\title[]{Object segmentation in the wild with foundation models: application to vision assisted neuro-prostheses for upper limbs}
	
	\author[1]{\fnm{Bolutife} \sur{Atoki}}\email{bolutife-oluwabunmi.atoki@etu.u-bordeaux.fr}
	
	\author[1]{\fnm{Jenny} \sur{Benois-Pineau}}\email{jenny.benois-pineau@u-bordeaux.fr}
	%equalcont{These authors contributed equally to this work.}
	
	\author*[2]{\fnm{Renaud} \sur{P\'eteri}}\email{renaud.peteri@univ-lr.fr}
	\author[1]{\fnm{Fabien} \sur{Baldacci}}\email{fabien.baldacci@u-bordeaux.fr}
	\author[3]{\fnm{Aymar} \sur{de Rugy}}\email{aymar.derugy@u-bordeaux.fr}
	
	\affil[1]{\orgdiv{LaBRI, CNRS}, \orgname{Univ. Bordeaux}, \orgaddress{\street{UMR 5800}, \city{Talence}, \postcode{F-33400}, \country{France}}}
	\affil[2]{\orgdiv{MIA}, \orgname{Univ. La Rochelle}, \orgaddress{\city{La Rochelle}, \postcode{F-17042},  \country{France}}}
	\affil[3]{\orgdiv{INCIA,CNRS}, \orgname{UMR 5287}, \orgaddress{\city{Bordeaux}, \postcode{F-33400}, \country{France}}}

	
	\abstract{In this work, we address the problem of semantic object segmentation using foundation models. We investigate whether foundation models, trained on a large number and variety of objects, can perform object segmentation without fine-tuning on specific images containing everyday objects, but in highly cluttered visual scenes. The ''in the wild'' context is driven by the target application of vision guided upper limb neuroprostheses. We propose a method for generating prompts based on gaze fixations to guide the Segment Anything Model (SAM) in our segmentation scenario, and fine-tune it on egocentric visual data. Evaluation results of our approach show an improvement of the IoU segmentation quality metric by up to 0.51 points on real-world challenging data of Grasping-in-the-Wild corpus which is made available on the \href{https://universe.roboflow.com/iwrist/grasping-in-the-wild}{RoboFlow} Platform\footnote{\url{https://universe.roboflow.com/iwrist/grasping-in-the-wild}}.} % (\url{https://universe.roboflow.com/iwrist/grasping-in-the-wild})


	\maketitle
	% Main text
	\section{Introduction}
	\label{sec:intro}
	Object detection and segmentation is a widely studied problem in image analysis, with many application areas including: surveillance \cite{8296499} or the detection of cars, cyclists and traffic signs for traffic control \cite{DBLP:conf/cvpr/ChenD022}. In robotic vision, \cite{10323166} implement dynamic obstacle detection and tracking for a small quadcopter using an RGB-D camera, through an ensemble of computationally efficient but low-precision models to achieve accurate object detection in real time. Assistive robotics is a field where methods for object detection, recognition, and segmentation are highly needed. 
	
	A subfield is the design of neuro-prostheses for arm amputees. While myoelectric signals from remaining muscles are typically used to control the prosthetic wrist and hand, computer-vision from camera worn on the prosthesis are increasingly proposed to assist the control, such as for the selection and control of the appropriate grasp type \cite{DBLP:journals/tcyb/ZhongHL22}. For above-elbow amputation, where remaining muscles are not sufficient to control the multiple lost degrees of freedom, object-pose gathered though computer vision could be even more critical. Indeed, we showed recently that Artificial Neural Networks (ANN) trained on natural arm movements predict distal joints so well that arm amputees could reach as with a natural arm in a virtual environment  \cite{segas_intuitive_2023}. Yet, the pose of the object of interest is a mandatory input for this type of ANN-based control. When using a camera on glasses with eye tracking capabilities, we showed in \cite{gonzalez-diaz_perceptually-guided_2019} that gaze could be leveraged to ensure detection of the correct object. As for the object pose, the state-of-the-art 6D pose estimator developed in \cite{wang_densefusion_2019} could be adapted to our needs, as indicated next and described in this contribution.  The method requires an object mask in the video frame or image scene, it is therefore necessary to accurately segment the object to be grasped. 
	Object segmentation involves identifying objects within images to precisely obtain object boundaries at the pixel level \cite{DBLP:journals/pami/ChenPKMY18}. Foundation models \cite{DBLP:journals/corr/abs-2108-07258} such as \texttt{Detectron2} \cite{wu2019detectron2} and \texttt{Segment Anything Model} (SAM) \cite{Kirillov_2023_ICCV} are trained on a large amount of data and are adaptable to a wide range of tasks. These models allow for high quality segmentation masks and precise boundaries with respect to human perception \cite{Kirillov_2023_ICCV}.
	Nevertheless, for the segmentation of specific objects, foundation models like SAM need guidance, i.e. "prompts" \cite{Kirillov_2023_ICCV}. This means that some indicators of the object of interest in the scene are needed. For applications in real-world scenarios of neuroprosthesis control, visual scenes are cluttered, as can be seen in the images of the dataset we recorded for this purpose. Therefore, we propose a method for generating prompts for object segmentation based on the prosthesis wearer's gaze fixations, recorded with a head-mounted device that integrates the eye trackers, and adapt the segmentation foundation model.\\
	
	The contributions of this paper are thus follows:  
	\begin{itemize}
		\item The use of gaze fixation information as a prompt for foundation model in an object segmentation task.% points available due to the vision-guided prosthesis scenario to obtain masks.
		\item Annotation of binary masks for the Grasping-in-the-Wild  (GITW) egocentric video dataset.
		\item Evaluation of the need for fine-tuning a foundation model for real-world cluttered video data in an object grasping scenario.
		\item Adding contextual information to the segmentation process and evaluating its impact on the accuracy of the predicted masks.
	\end{itemize}
	
	The remainder of the paper is organized as follows: in section \ref{sec:lit-review}, we give a brief overview of popular segmentation models and justify our choice of a baseline model. In section \ref{sec:methodology}, we present our approach for adapting the foundation model for object segmentation in a vision-guided neuroprosthesis scenario. In section \ref{sec:results}, we describe our real-world data, the evaluation metrics used, and the results obtained. Section \ref{sec:conclusion} concludes this work and outlines its perspectives.
	
	\section{Literature Review}
	\label{sec:lit-review}
	
	Image segmentation is a research problem with a long history. Initially, it meant partitioning the image plane into a set of non-overlapping regions satisfying a low-level homogeneity criterion as chrominance or shape \cite{ChasseryG84}. This method uses an iterative process that first divides the image into regions based on \textit{a priori} information (shape or uniformity), and then updates each region obtained by aggregating connected points using colour similarity measured by Euclidean distance and spatial cohesion. To extract objects with different homogeneous regions inside, external cues were included, e.g. in video scenes, the regions animated by a homogeneous motion would represent semantic objects or the background. Homogeneous regions are obtained based on the detection and extraction of contrasting contours and a region growing of these contours using a quad-tree decomposition. For video frames, motion vectors are computed to observe the movement of regions across successive frames and used to predict the new positions of these regions in subsequent frames. Homogeneous object motion over a temporal window is then used to merge adjacent regions.
	
	With the advent of supervised machine learning approaches, image segmentation has taken a step towards semantic segmentation, with the introduction of semantic homogeneity of sets of neighbouring regions in the image. Image segments are obtained from known objects in the image using segmentation algorithms including grid and graph-based segmentation. Colour and texture features are extracted from these segments and concatenated to form a feature vector describing the segment. These vectors and their corresponding labels are provided as input to a support vector classifier for training and evaluation.
	
	However, the real breakthrough in semantic object segmentation was achieved by using deep convolutional neural network architectures trained on ground truth segmentation masks. For example, \cite{DBLP:conf/nips/CiresanGGS12} use convolutional features as input to a pixel-wise classifier for segmenting neural structures in electron microscopy (EM) images. 
	
	Another approach consists in the use of overlapping patches of the image as input to convolutional networks like AlexNet \cite{DBLP:conf/nips/KrizhevskySH12}, VGG \cite{DBLP:journals/corr/SimonyanZ14a}, GoogLeNet \cite{DBLP:conf/cvpr/SzegedyLJSRAEVR15} which have their fully connected layers replaced with $1 \times 1$ convolutional layers having filter dimensions corresponding to the number of object classes and background. The obtained downsampled feature/heatmaps are stitched together in the same order as their patches to form the complete heatmap representing the image. To combine coarse semantic information with fine appearance information, features from deep layers are concatenated with those from shallow layers and upsampled to match the input image size. The result of this is an image whose resolution matches the one of the input image, having dimensions corresponding to the number of object classes. \cite{DBLP:conf/bildmed/Ronneberger17} implement a novel symmetric \textit{U-shaped} encoder-decoder network, utilizing high-resolution features from the encoder network combined with up-sampled features in the decoder network to help the model, obtain precise masks. As a result, the encoder network has a high number of feature channels, thus allowing the network to propagate contextual information to its high-resolution layers. Similar to the skip connection and symmetry of the U-Net, \cite{DBLP:conf/bildmed/Ronneberger17, DBLP:journals/pami/BadrinarayananK17} designed a network for image segmentation that substitutes learnable up-sampling layers with non-linear up-sampling by utilizing the max-pooling indices from the corresponding encoder layer reducing model parameters (from 134 M to 14.7 M) and improving boundary delineation. \cite{DBLP:conf/eccv/ChenZPSA18} take skip connections, encoder-decoder architecture, and are reducing computation parameters even further with the introduction of atrous-separable convolution and the simplistic yet effective decoder implemented in the Deeplabv3+ model. The atrous-separable convolution combines the benefit of atrous convolution which allows obtaining multi-scale features by adjusting the convolution kernel's field-of-view and that of separable convolution allows for much less computation by splitting the convolution into depth-wise and point-wise with a $1\times 1$  kernel. This  yields to a reduction in the number of multiplication operations during convolution while keeping the same outputs.

	Computer vision tasks like object detection, classification, and even segmentation since the widespread adoption of CNNs and other Deep learning architectures relied on applying these models trained on labelled datasets like ImageNet \cite{DBLP:conf/cvpr/DengDSLL009}, MS Coco \cite{DBLP:conf/eccv/LinMBHPRDZ14} as feature extractors \cite{yolov5}, by fine-tuning on custom dataset \cite{DBLP:journals/sncs/VenkateshAGYP24}, and in some cases with zero-shot application \cite{DBLP:conf/mm/LiuZGZ23}. This trend of adapting models trained on these widely accepted datasets led to the development of much deeper, larger networks trained on a variety and vast amount of data. These models were first defined as Foundation Models by \cite{bommasani2021opportunities} and refer to models trained on broad, large-scale data that can be adapted to perform a range of downstream tasks including those previously mentioned. \cite{DBLP:journals/corr/abs-2307-13721} refer to foundation models as models pre-trained on large-scale, extensive and diverse data which can be used as the base for various computer vision downstream tasks and generalize well across them. With the emergence and development of these foundation models, research on their application at the zero-shot level and adaptation -- fine-tuning -- methods have been performed in different computer vision tasks, particularly in the aspect of pixel-wise segmentation of images. \cite{DBLP:conf/cvpr/ZhangDS24} investigate the use of Foundation models SAM \cite{Kirillov_2023_ICCV}, DINOv2 \cite{DBLP:journals/tmlr/OquabDMVSKFHMEA24} in low-resource scenarios having scarcity of data, precise attention to detail, and a significant domain shift from natural images to the specialized domain of interest with a zero-shot approach. It was quickly observed in terms of performance metric values that the foundation models in a zero-shot approach were not appropriate for these tasks. The authors proposed three methods to overcome this - implementing generative models to create more dataset images, reducing the patch-size used for tokenisation to capture finer image details, and learning attention maps that highlight important regions for each low resource domain. Mask2Former \cite{DBLP:conf/cvpr/ChengMSKG22} is a Foundation model that provides a unified framework for semantic, instance, and panoptic segmentation. The model adopts the same architecture as MaskFormer \cite{DBLP:conf/nips/ChengSK21} - a feature extractor block which encodes the input image into a feature representation. Here, two networks are made available: ResNet \cite{DBLP:conf/cvpr/HeZRS16} and SwinV2 \cite{DBLP:conf/iccv/LiuL00W0LG21}. It also includes an up-sampling pixel decoder, and a transformer decoder, with changes to incorporate multi-level features in the transformer block by sharing activation maps from each decoder pyramid layer to all transformer layers. The transformer decoder utilizes learnable query features which are initially initialized to zero and updated during training, to predict a temporary mask (through the projection of the query features to the image resolution) which is used for masked attention by restricting the processing of the image features from the "encoder" block, to the foreground region of the temporary mask. This process happens per layer and refinement occurs by the progression through all layers in the "Transformer Decoder" block. This results in better performance than the MaskFormer \cite{DBLP:conf/nips/ChengSK21} model on segmentation tasks with much fewer training resources, as well as a generalized network for instance, semantic, and panoptic segmentation competitive with state-of-the-art methods. Detectron2 \cite{wu2019detectron2} is a foundation model with training datasets including Pascal VOC, ADE20k Scene Parsing, cityscapes, coco, LVIS, and PanopticFPN that performs downstream tasks of object detection, instance, and semantic segmentation. It uses backbone feature extractor networks \cite{DBLP:conf/cvpr/HeZRS16, DBLP:conf/cvpr/XieGDTH17} for convolutional features, fed to the region proposal network for object regions, which are pooled and passed into respective prediction heads. The Detectron2 foundation model accepts images as input for its tasks, while Segment Anything (SAM). \cite{Kirillov_2023_ICCV} additionally accepts optional prompts in the form of object points, object bounding box, low-resolution object mask,... These additional prompts improve the quality of obtained masks. In the vision-guided prosthesis scenario, the gaze fixations of the amputee are measured and generally located on the object to be grasped. Therefore, it is natural to resort to the SAM segmentation model, using these gaze fixations as the object points prompt, to improve segmentation accuracy in real-world cluttered scenes.
	
	\section{Methodology}
	\label{sec:methodology}
	In this section, we present our segmentation scenario along with proposed methods. 
	
	\subsection{Segmentation scenario}
	\label{subsec:seg_scenario}
	
\begin{figure}[tb]
		\centering
		\centerline{\includegraphics[width=7cm, height=7cm, keepaspectratio]{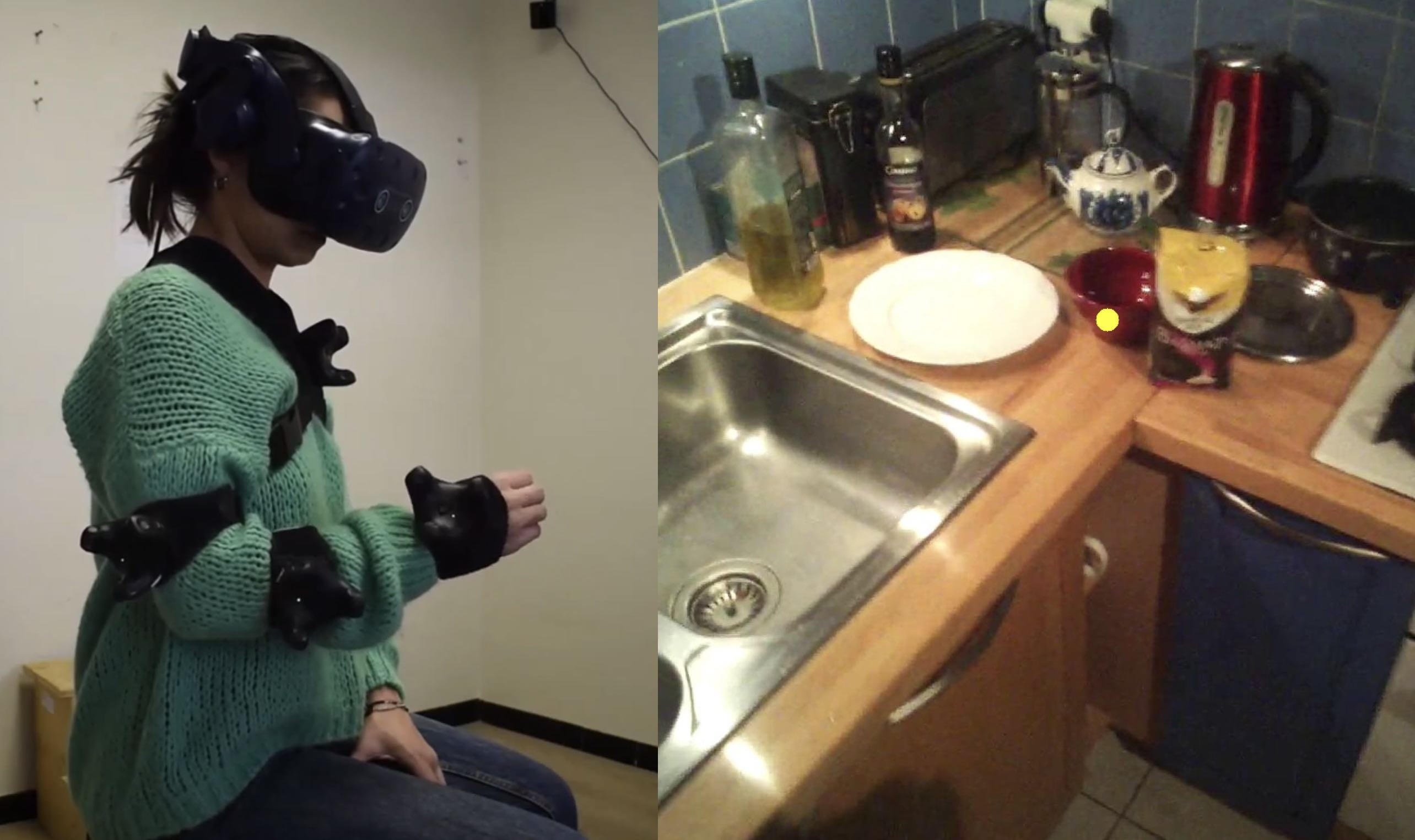}}
		\caption{Illustration of our segmentation scenario, inspired by \cite{Lento2024.01.30.577386}}
		\label{fig:gaze-point}
	\end{figure}

	In our scenario, the object of interest is indicated by the user's gaze before the reaching and grasping movement begins. Gaze fixations are a minimal piece of information that expresses the user's intention to grasp an object. Independently, it can be extended with the knowledge of the rules of foveal projection, into the image plane as in \cite{ObesoBGR22}. An example of a recorded gaze fixation on the object (bright point) is given in fig.\ref{fig:gaze-point}. This is how subjects look at the objects they want to grasp. Objects present in the scene should be segmented before the grasp. In real-world scenarios, the visual scenes are very cluttered, as shown in Figure \ref{fig:gaze-point}, right image. Therefore, we use the gaze fixation prompt and adapt this noisy information for its efficient use as an input to the foundation segmentation model. In the following, we briefly review the foundation model we have chosen and describe in detail how we adapt gaze fixations to drive it.
	
	\subsection{The Segment Anything Model}
	\label{subsec:sam}
\begin{figure}[h!]
		\centering
		\includegraphics[width=10cm, height=10cm,keepaspectratio]{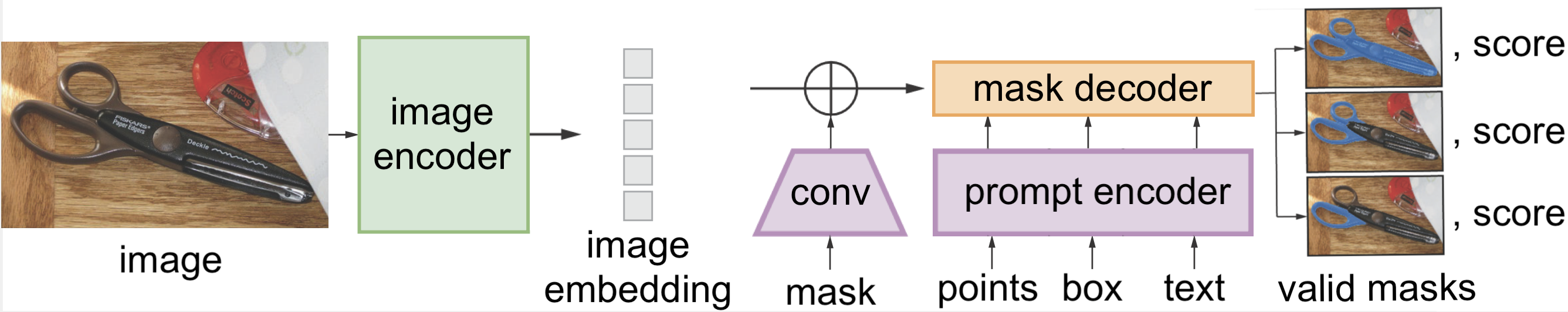}
		\caption{Segment Anything Architecture}
		\label{fig:sam-arch}
	\end{figure}

	The Segment Anything Foundation Model (SAM) \cite{Kirillov_2023_ICCV} consists of three main blocks, as illustrated in figure \ref{fig:sam-arch}. 
	\begin{enumerate}
		\item \textbf{Image Encoder:}\\
		Inputs are images of dimensions $C\times H\times W$ ($C$, $H$ and $W$ being respectively the number of channels, the height and the width of the input image). This step  uses a Vision Transformer (ViT) \cite{DBLP:conf/iclr/DosovitskiyB0WZ21} to generate a $16 \times$ down-scaled feature embedding of the input image (the "encoder" block in figure \ref{fig:sam-arch}). The generated embedding serves as part of the input to the mask decoder. 
		
		\item \textbf{Prompt Encoder:}\\
		The prompt encoder receives optional prompt inputs in the form of image points (foreground or background) and corresponding binary labels. Other options are the bounding box coordinates of the object of interest, an initial low-resolution mask of the object, or a text description of the object (yet to be implemented by Meta AI). In our approach, we use only image points as additional prompt. For a point input, the encoder returns the sum of the positional encoding of the point and a learned embedding that varies depending on whether the point is foreground (on the object) or background (off the object), see the "prompt encoder" block in figure \ref{fig:sam-arch}. 
		
		\item \textbf{Mask Decoder:}\\
		The Mask Decoder is a modified Transformer Decoder Block \cite{DBLP:conf/nips/VaswaniSPUJGKP17} that receives all available input embedding vectors, and uses bi-directional self-attention and cross-attention to update them. The embedding vectors are then upsampled and fed to an MLP, which predicts the probability of each pixel belonging to the foreground class, see the block ``mask decoder" in figure \ref{fig:sam-arch}. The reader can refer to \cite{Kirillov_2023_ICCV} for more details.
	\end{enumerate}
	Therefore, we need to adapt the information available in our scenario - the use of gaze fixation points - into foreground points for the model. In the following, we describe the adaptation steps from acquisition to projection, filtering, clustering and finally labelling. 
	
	\subsection{Extracting Prompts for SAM model}
	\label{subsec:gaze-points}
	The object segmentation information is captured using a device that provides a video sequence at 25 frames per second and is integrated with an eye tracker that captures gaze fixation information at a rate of 50Hz. Because the video frame rate is fixed at 25 Hz, the gaze points are interpolated  using spline interpolation to fit the number of frames\cite{799930}. This also reduces the amount of data to be processed.

	\subsubsection{Gaze Fixation Points}
	\label{subsec:gaze-points-projection}
\begin{figure}[h!]
		\begin{minipage}[c]{0.45\linewidth}
			\includegraphics[width=\linewidth]{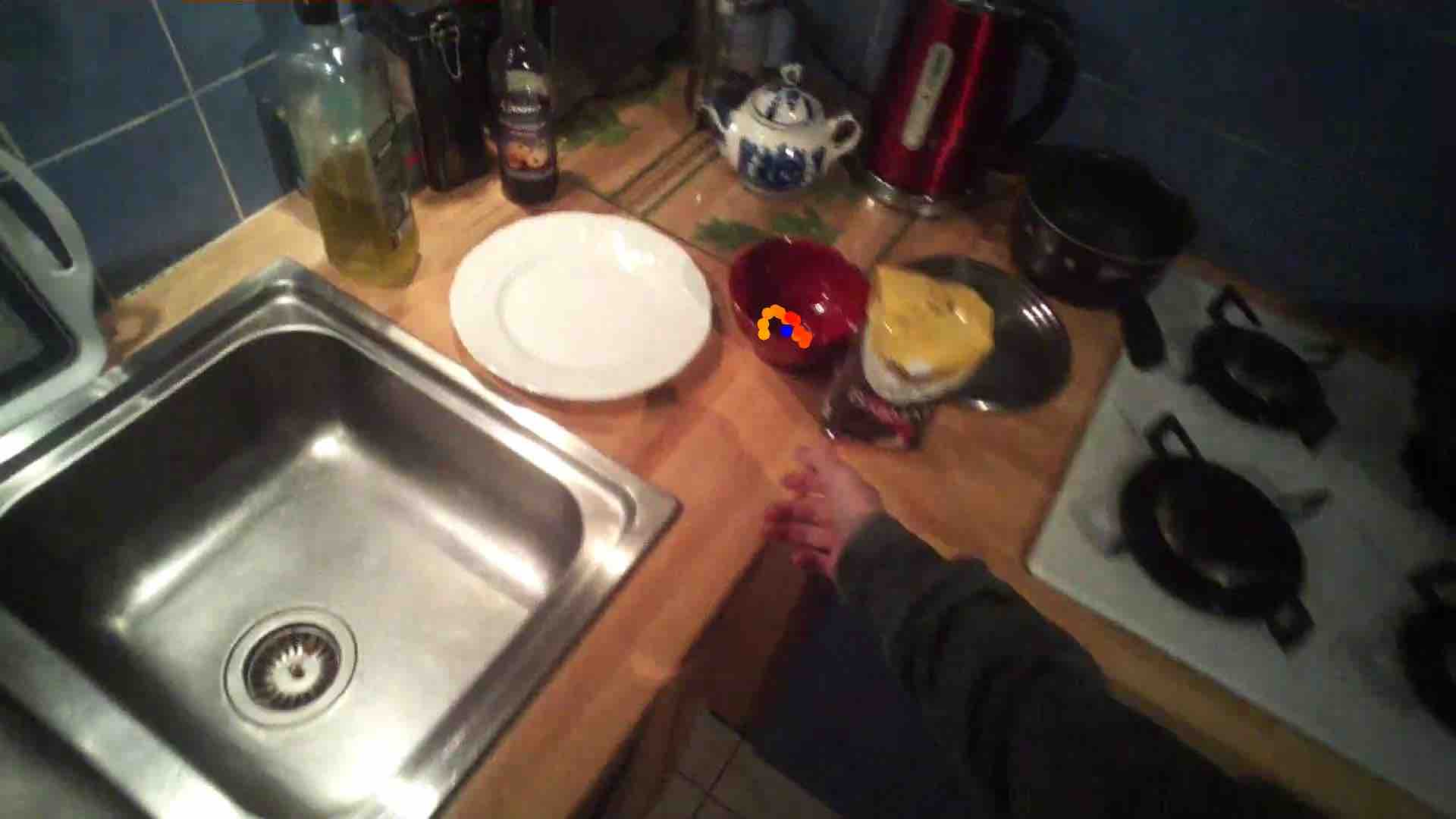}
			% \caption{a}
			\label{fig:proj-gp1}
		\end{minipage}
		\hfill
		\begin{minipage}[c]{0.45\linewidth}
			\includegraphics[width=\linewidth]{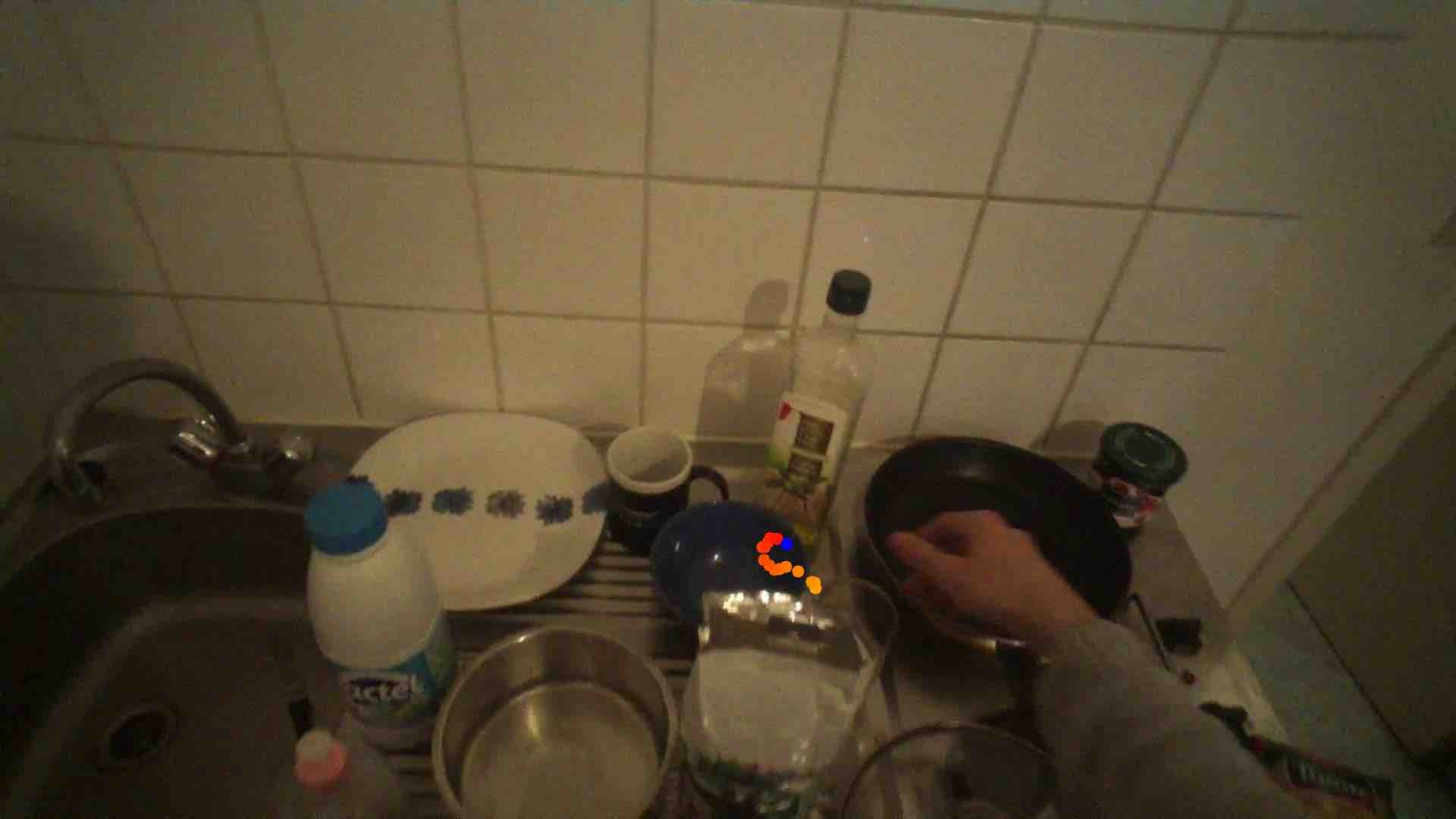}
			% \caption{b}
			\label{fig:proj-gp2}
		\end{minipage}
		\begin{minipage}[c]{0.45\linewidth}
			\includegraphics[width=\linewidth]{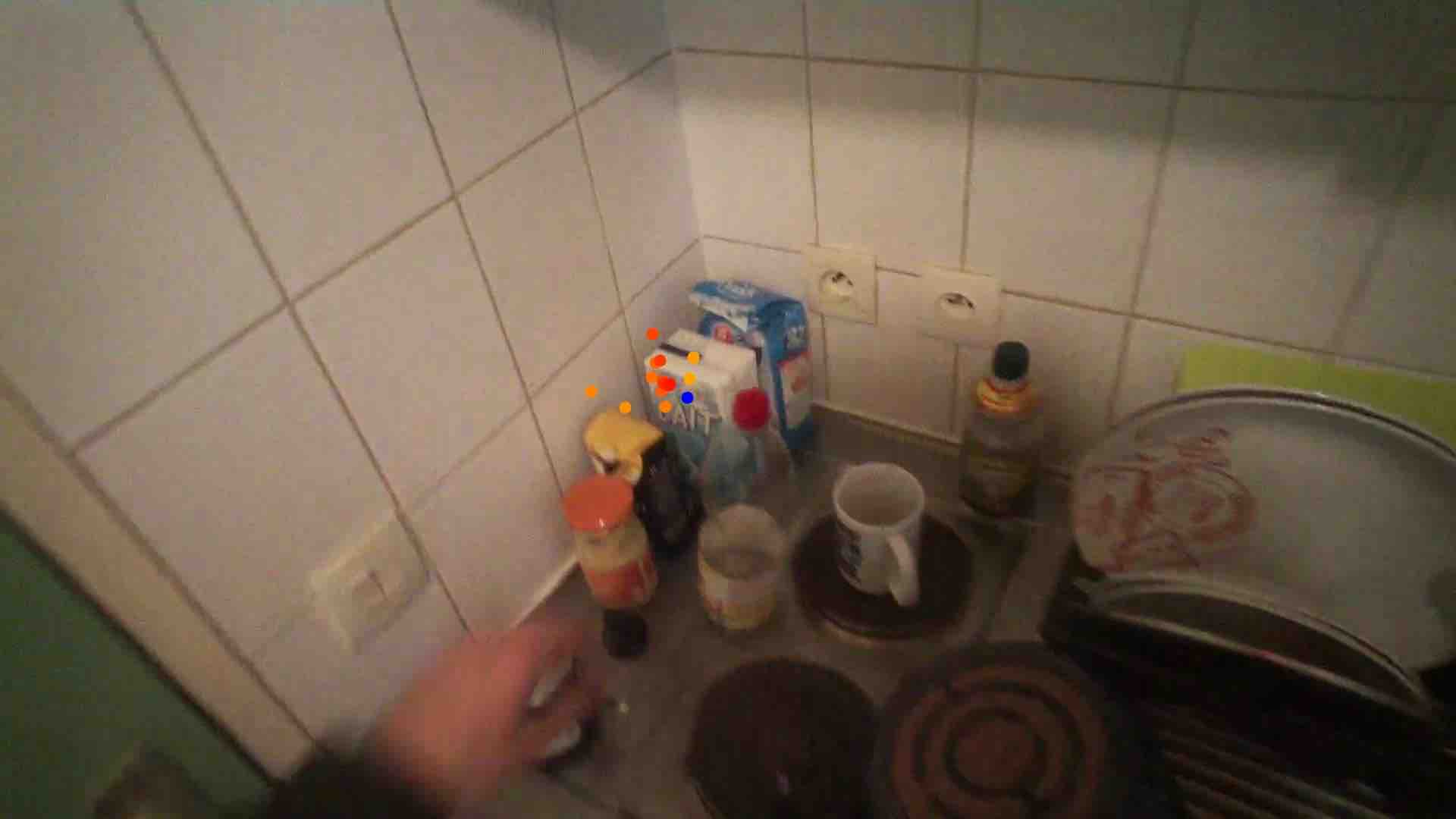}
			% \caption{c}
			\label{fig:proj-gp3}
		\end{minipage}
		\hfill
		\begin{minipage}[c]{0.45\linewidth}
			\includegraphics[width=\linewidth]{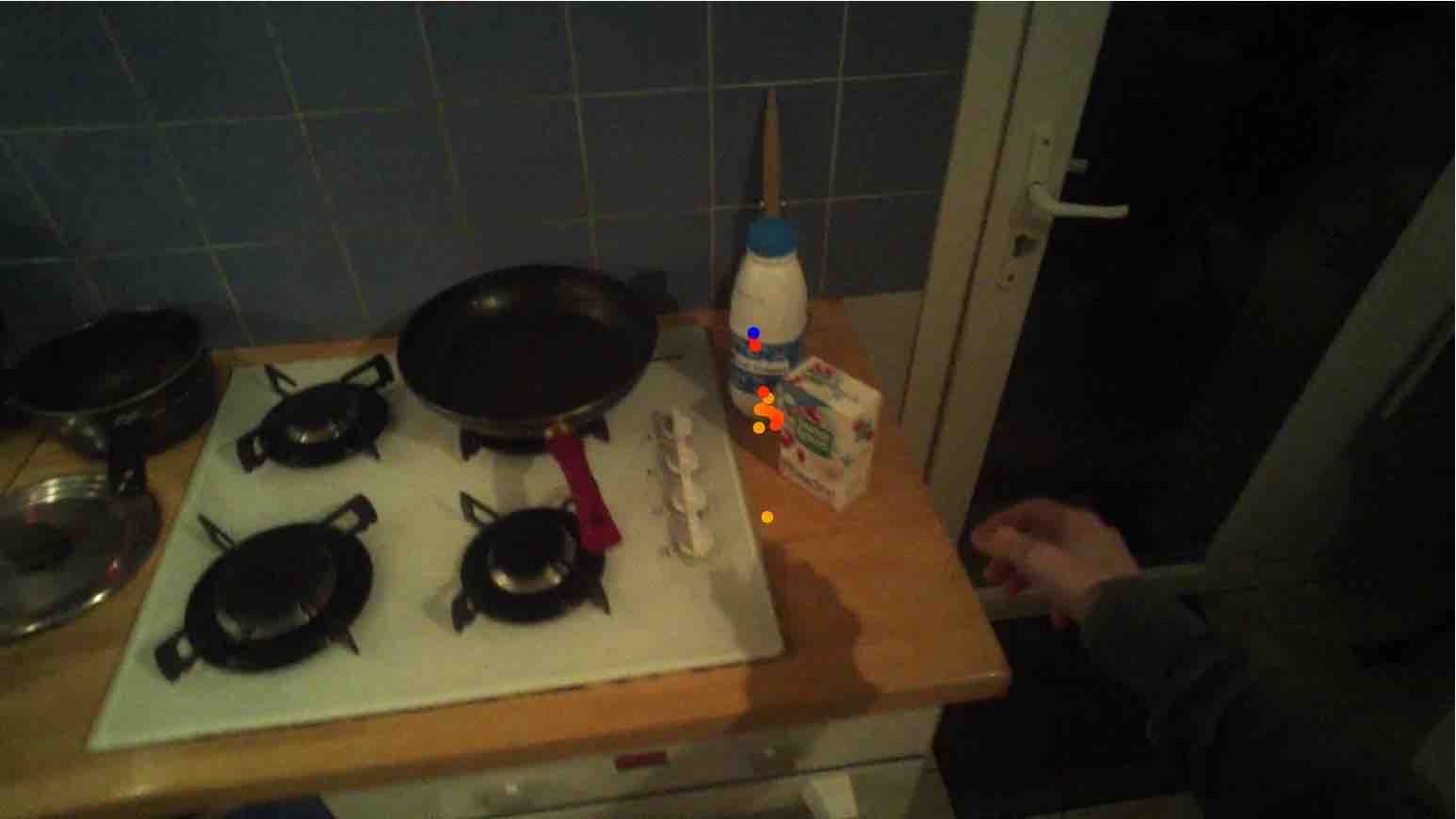}
			% \caption{d}
			\label{fig:proj-gp4}
		\end{minipage}
		
		\caption{Gaze points extracted from frames in the temporal window and projected onto the current frame. The most recent points are shown in red.}
		\label{fig:projected-gaze-points}
	\end{figure}

	The nature of the task being performed, lighting conditions or visual distractors in the scene can cause rapid changes in a subject's fixation point \cite{DBLP:conf/wacv/KadnerTHR23}.  This is called a visual saccade, which temporarily shifts the fixation away from the object of interest.
	Thus, relying on the single gaze point for the current frame to locate the object of interest could lead to errors. Therefore, a projection of previous gaze points onto the current frame is implemented, using homography estimation between successive frames. In the present work, the implementation is modified to reduce the propagation of errors across frames, since the homography model assumes that the scene contains planar surfaces and may not accurately capture more complex distortions in the video scene. The proposed solution uses a temporal window of $T$ frames to prevent this propagation by chaining the estimated homography matrices between frames in the temporal window by matrix multiplication. The final transformation is thus a composition of homographies as expressed in equation \ref{eqn:homography-chaining}:
	\begin{equation}
		\label{eqn:homography-chaining}
		\tilde{P}_t = H_1 \times H_2 \times \ldots \times H_{T-1} \times H_T \times P_{t-T+k}
	\end{equation}
	
	where $\tilde{P}_t$ = \((\tilde{x}, \tilde{y}, 1)_t\) are homogeneous gaze point coordinates at frame \(t\); \(H_n, n=1,2,...,T\) are the homography matrices between frames \(n\) and \(n+1\); \(T\) is the temporal window size, and $k= 0,...,T-1$. \\
	Note that in the current frame $t$, all gaze fixation points are considered: the projected ones $\tilde{P}_t$ as well as the current recorded gaze point $P_t$.\\
	Examples of gaze points extracted using a temporal window and projected onto the current frame are shown in figure \ref{fig:projected-gaze-points}.

	\subsubsection{Visual Saccades Elimination using Density Based Clustering}
	\label{subsec:clustering}
	Visual saccades represent outliers in the distribution of gaze points. To eliminate outliers, we perform a clustering of the points using a density-based method, as proposed in \cite{DBLP:journals/jimaging/FejerNBSRD22}. The aim is to identify the population of points that are close to each other and thus located on the object of interest. Density Based Clustering (DBSCAN) is a suitable algorithm for this purpose. DBSCAN \cite{silverman2018density} is a well-known non-parametric clustering algorithm that does not require a target number of clusters to be defined and can handle clusters of non-spherical shape. This is appropriate for our case because humans perform micro-saccades and never foveate the same spatial location over time. The algorithm is controlled by two parameters: the minimum number of points ($minPts$) and the radius of the point neighbourhood ($\epsilon$ value). The $minPts$ parameter specifies the high-density areas, and in our case its upper bound is the temporal window size $T$. The $\epsilon$ parameter determines the distance threshold for points to be considered neighbours. It depends on the size of the microsaccades while the subject maintains fixation on the object.
	
\begin{figure}[h!]
		\begin{minipage}[c]{0.45\linewidth}
			%\captionsetup{justification=centering, format=plain, width=0.15\linewidth} 
			\includegraphics[width=\linewidth]{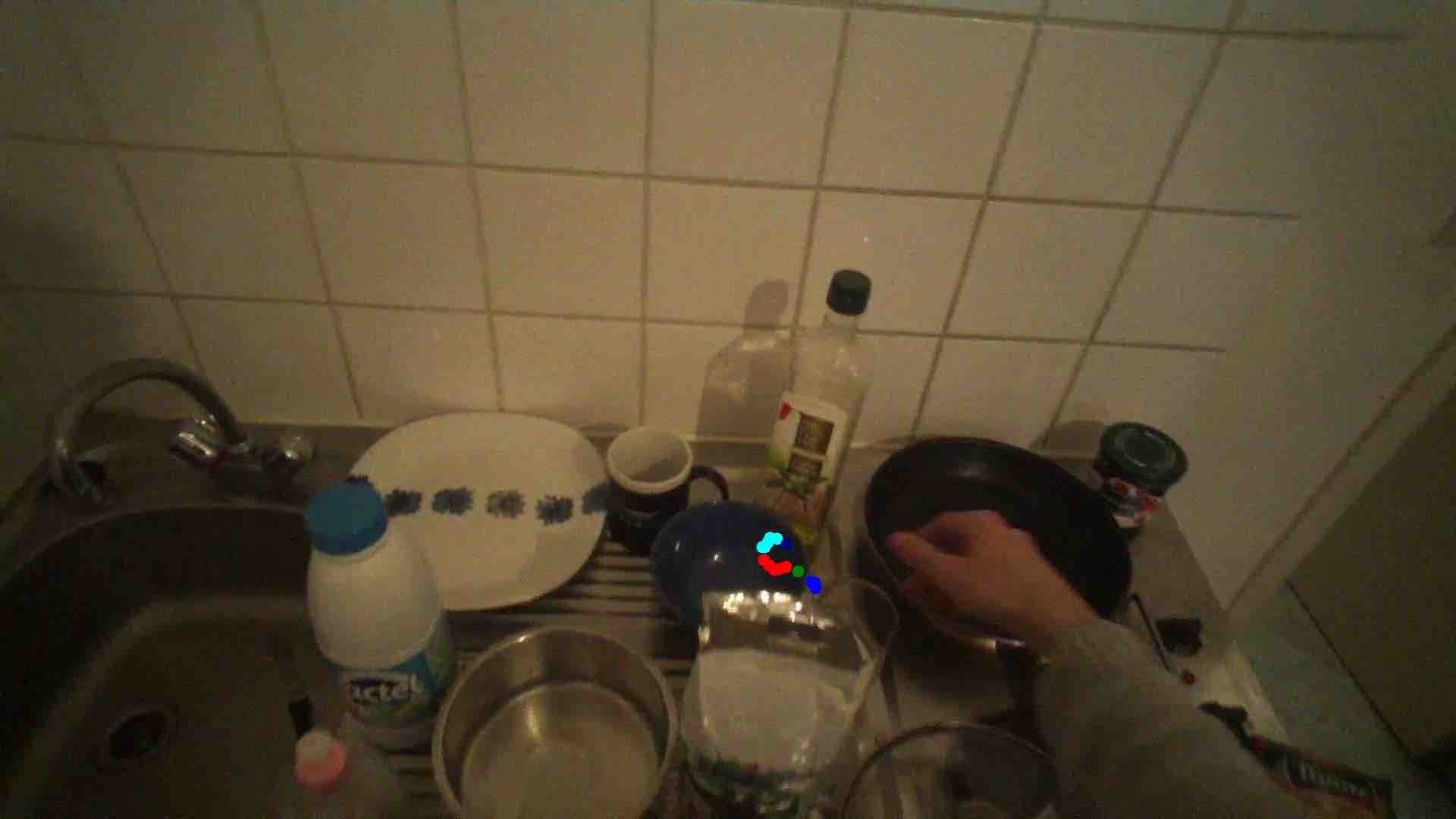}
			% \centering
			\centerline{Gaze point clusters}
			\label{fig:all-clusters-Bowl}
		\end{minipage}
		\hfill
		\begin{minipage}[c]{0.45\linewidth}
			\includegraphics[width=\linewidth]{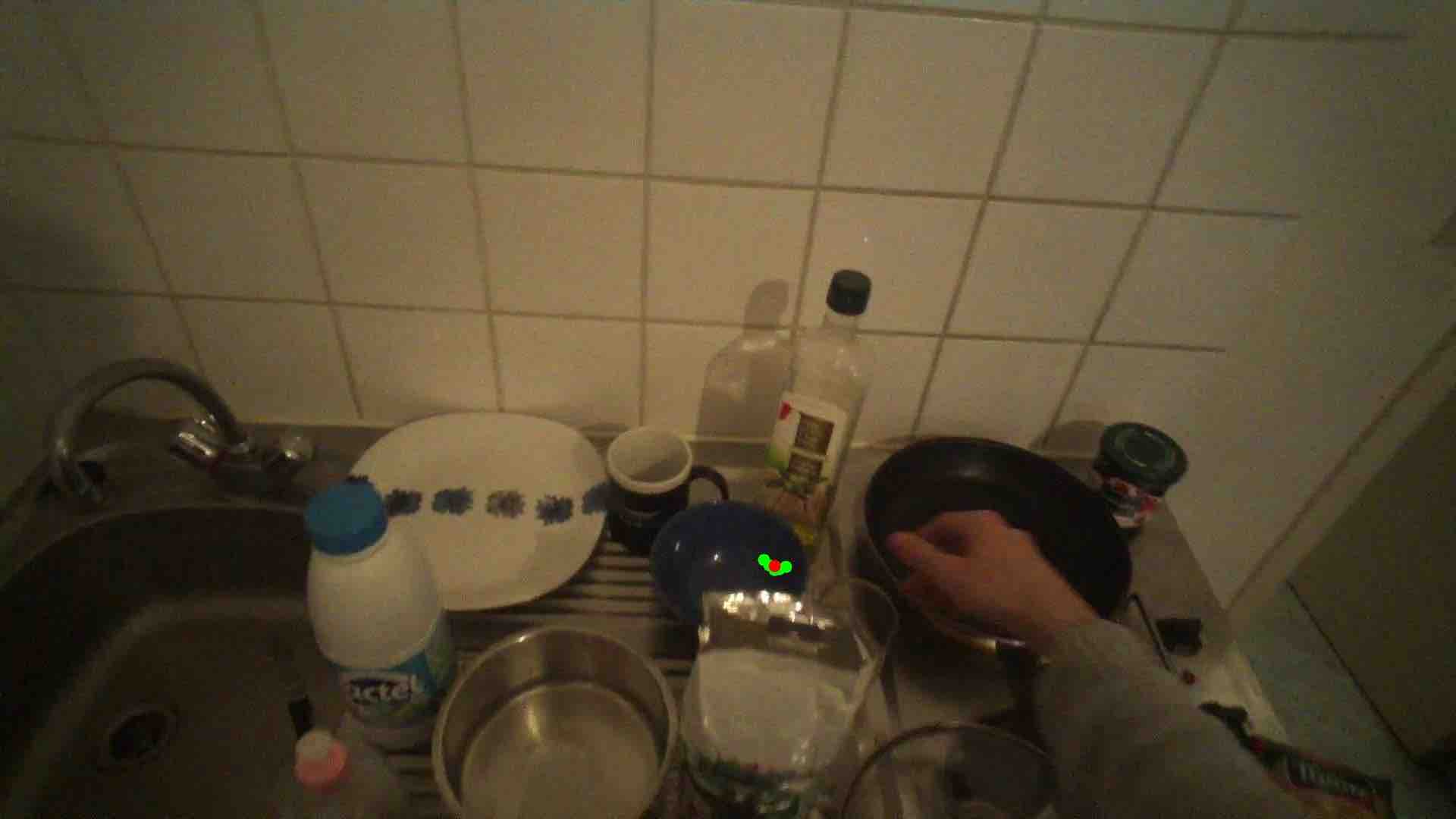}
			\centerline{Largest cluster selected.}
			\label{fig:largest-cluster-Bowl}
		\end{minipage}
		\begin{minipage}[c]{0.45\linewidth}
			\includegraphics[width=\linewidth]{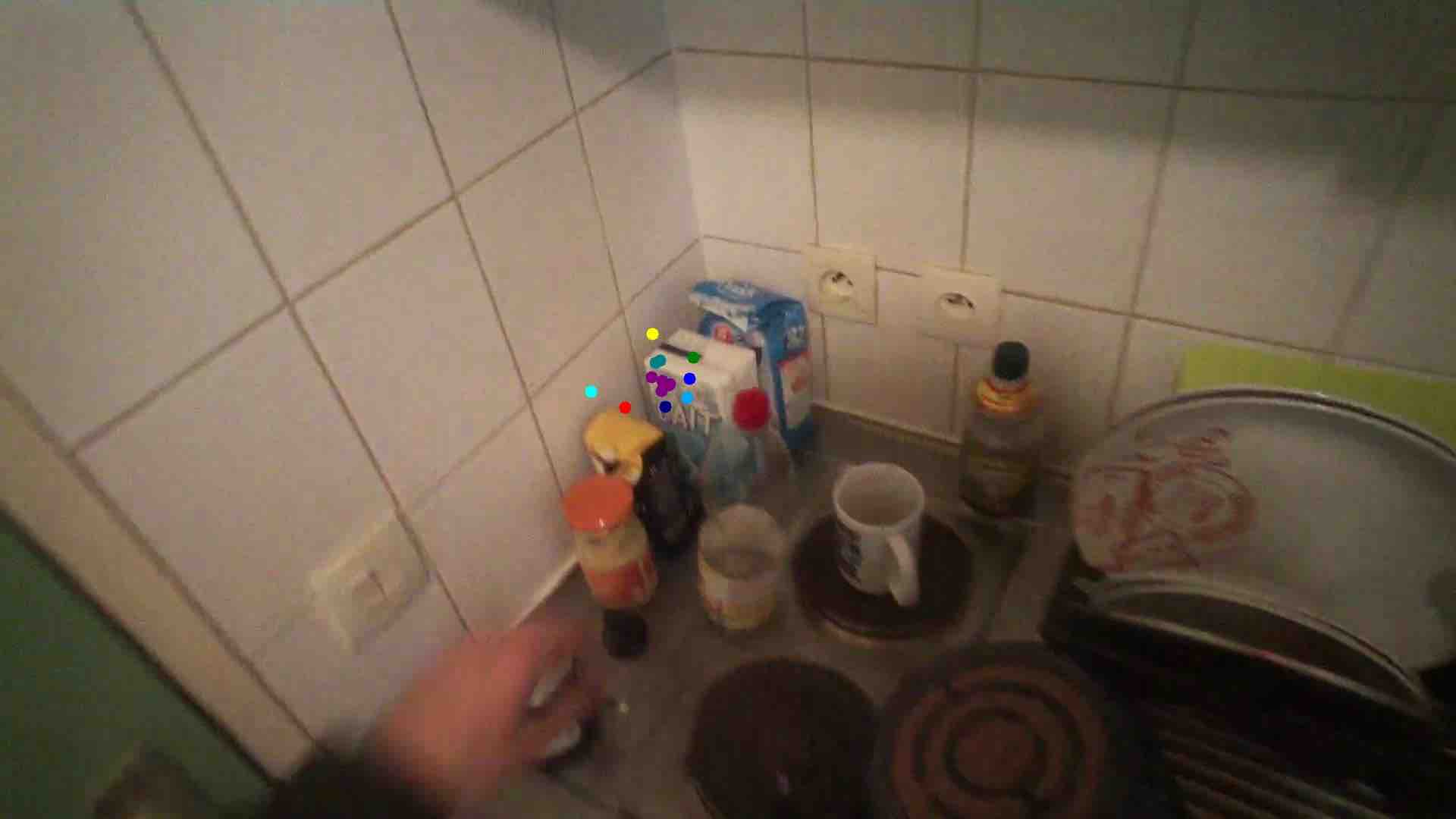}
			\centerline{Gaze point clusters}
			\label{fig:all-clusters-Milk}
		\end{minipage}
		\hfill
		\begin{minipage}[c]{0.45\linewidth}
			\includegraphics[width=\linewidth]{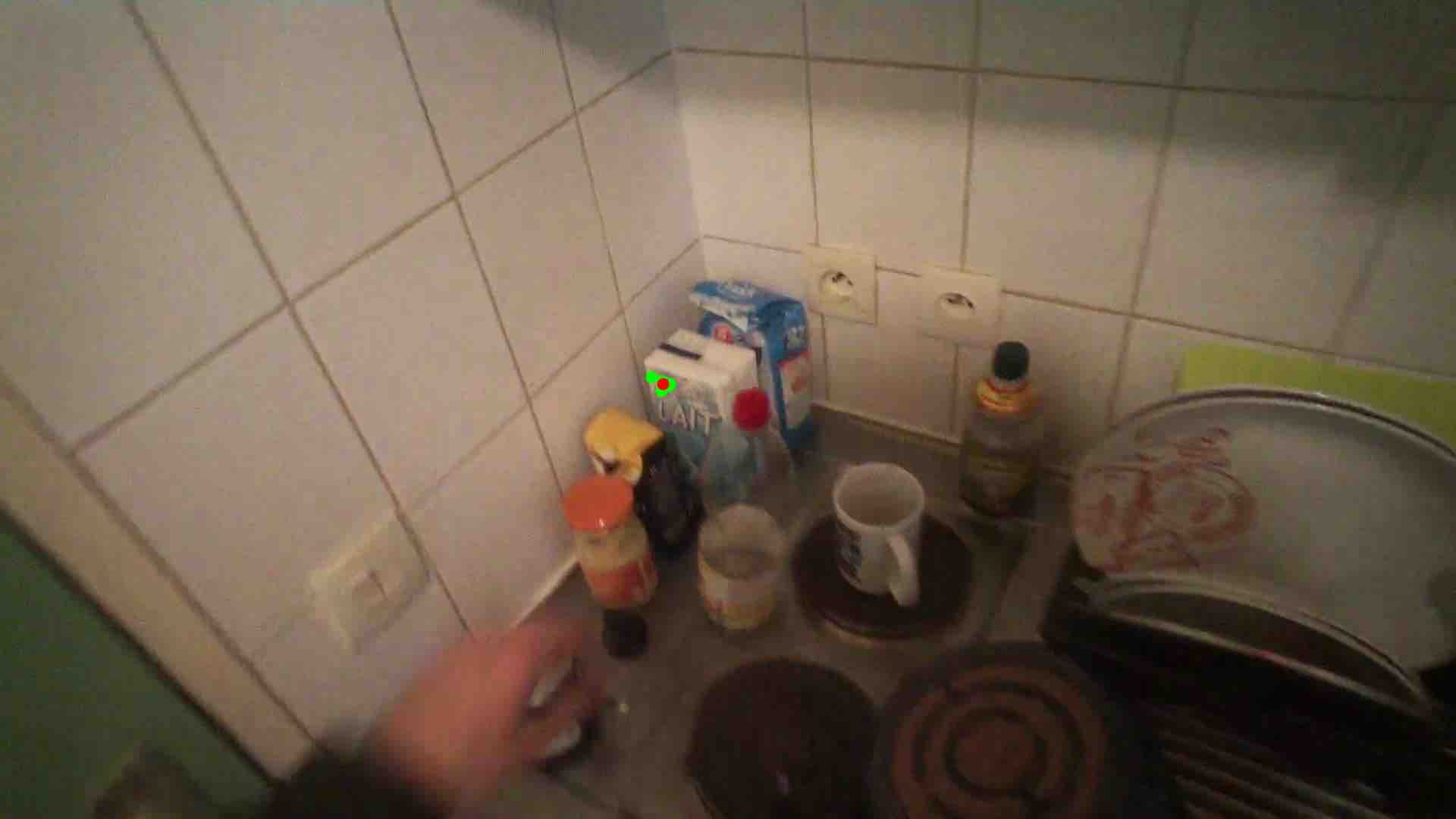}
			\centerline{Largest cluster selected.}
			\label{fig:largest-cluster-Milk}
		\end{minipage}
		
		\caption{Outlier detection and elimination process using density-based clustering}
		\label{fig:dbscan}
	\end{figure}

	According to analysis of the literature made in  \cite{10.1167/8.14.21}, the average microsaccade magnitude is \textbf around 0.5 visual degrees. Hence, taking into account the video acquisition conditions and distance from the glasses-worn eye-tracker, we set $\epsilon = 1.4$ pixels in a video frame.\\
	Examples of this outlier detection and elimination process are shown in figure \ref{fig:dbscan}.
	
	\newpage
	\subsection{Using SAM to segment objects in the grasping scenario}
	\label{subsec:sam-application-approaches}
	The SAM foundation model chosen is applied to the grasping scenario for three considered approaches.
	\begin{enumerate}
		\item Application of the pre-trained model to the real-world cluttered GITW dataset\footnote{\url{https://universe.roboflow.com/iwrist/grasping-in-the-wild}}.
		\item Fine-tuning of the pre-trained model with ground truth binary masks and applying it to the GITW dataset.
		\item Fine-tuning of the pre-trained model with automatically computed fuzzy masks and applying it to the GITW dataset.
	\end{enumerate}
	
	An explanation and justification of the proposed approaches is provided in the following subsections. 
	
	\subsubsection{Application of pre-trained SAM to the GITW Dataset}
	\label{subsubsec:sam-application-pre-trained}
\begin{figure}[h!]
		\centering
		\includegraphics[width=0.7\textwidth]{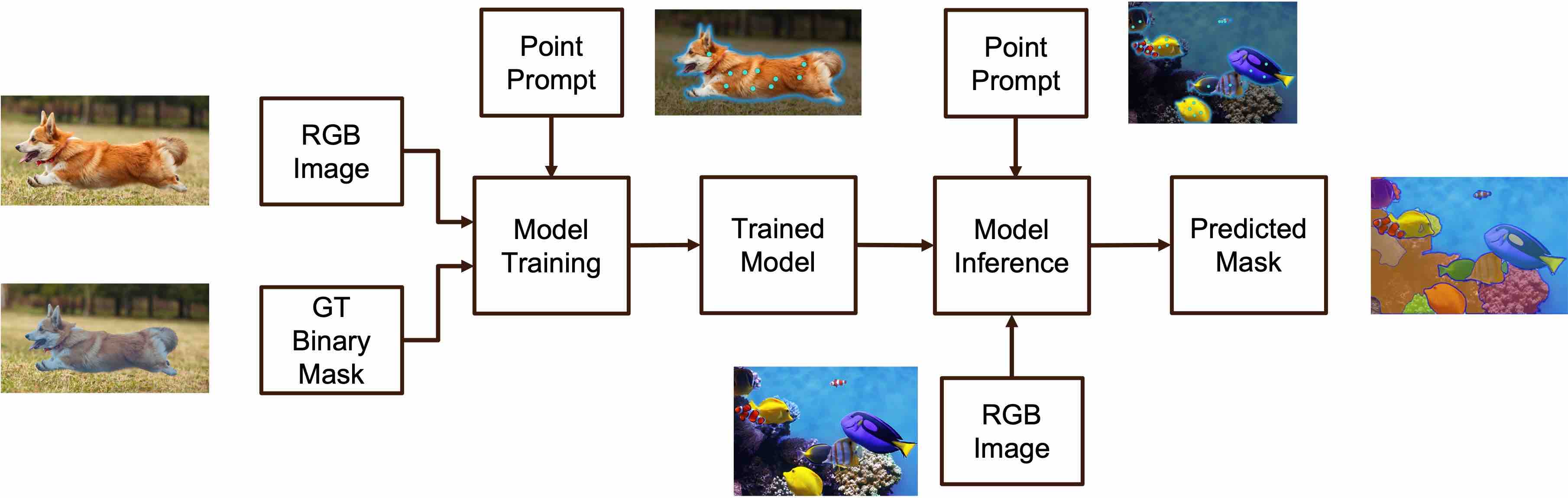}
		\caption{Application of SAM \cite{Kirillov_2023_ICCV} without fine-tuning}
		\label{fig:sam-application-pretrained}
	\end{figure}

	The paper introducing SAM \cite{Kirillov_2023_ICCV} claims that such a model can segment natural objects. Therefore, our first approach is the application of SAM to natural data. This pre-trained model is applied directly to the GITW dataset and used to extract object segmentation masks. This can be modeled as: 
	\begin{equation}
		\label{eqn:WithoutFine-Tunning}
		\text{M}(\text{D}_1) \rightarrow \text{D}_{2(test)}
	\end{equation}
	
	where $\text{M}$ is the Foundation Model, $\text{D}_1$ is the Large Scale dataset, $\text{D}_{2(test)}$ is the target test set, the test set of \texttt{GITW} in our case. 
	This process is represented in figure \ref{fig:sam-application-pretrained}.
	
	\subsubsection{Fine-tuning of SAM with binary masks of the GITW Dataset}
	\label{subsubsec:sam-application-bin-mask}
\begin{figure}[h!]
		\centering
		\includegraphics[width=0.9\textwidth]{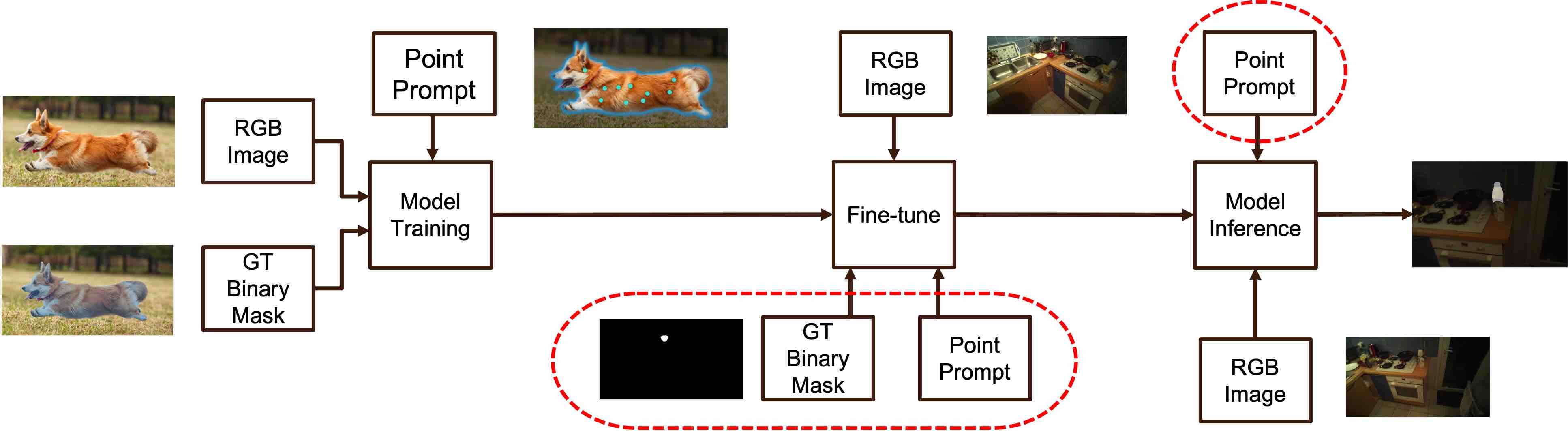}
		\caption{Application of SAM \cite{Kirillov_2023_ICCV} fine-tuned on considered dataset's binary masks }
		\label{fig:sam-application-finetuned}
	\end{figure}

\begin{figure}[h!]
		
		\begin{minipage}[c]{0.45\linewidth}
			\includegraphics[width=\linewidth]{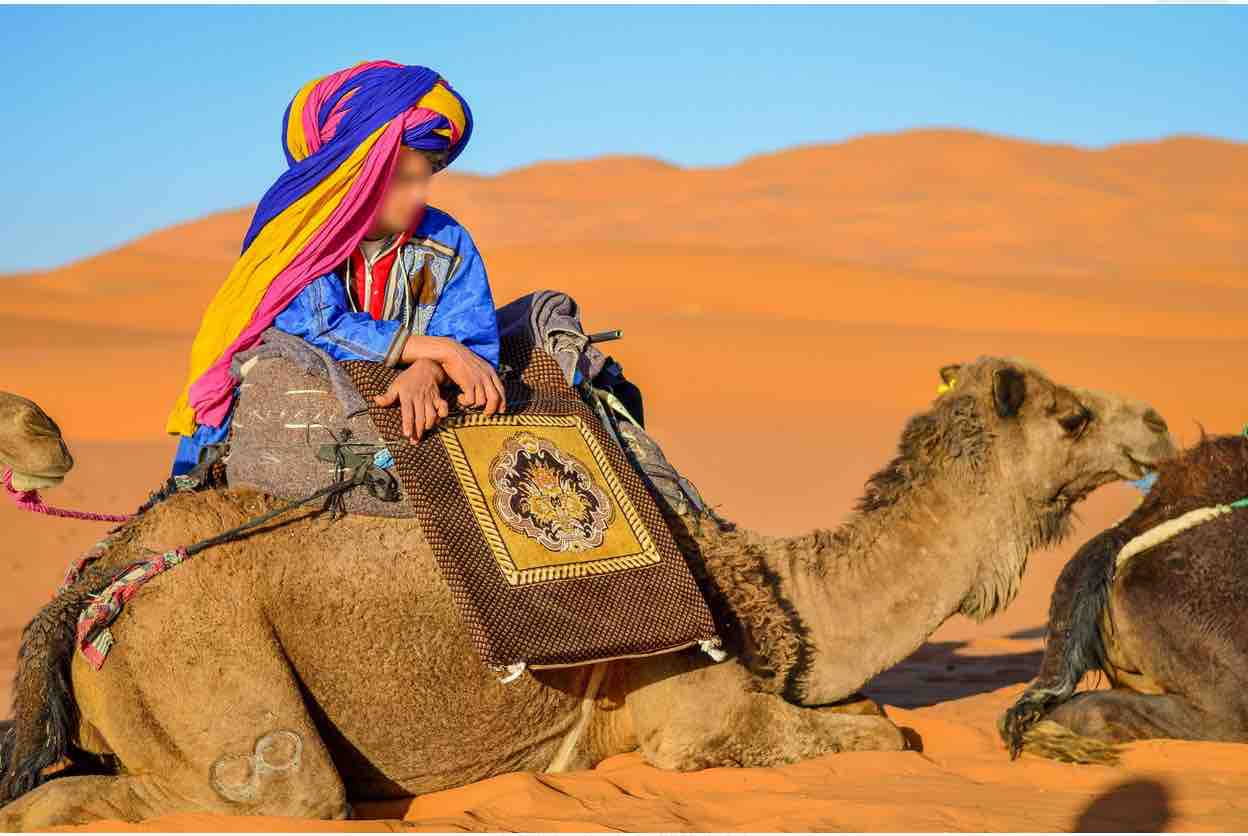}
			\centering{RGB image}
			\label{fig:sam-dataset-sample1}
		\end{minipage}
		\hfill
		\begin{minipage}[c]{0.45\linewidth}
			\includegraphics[width=\linewidth]{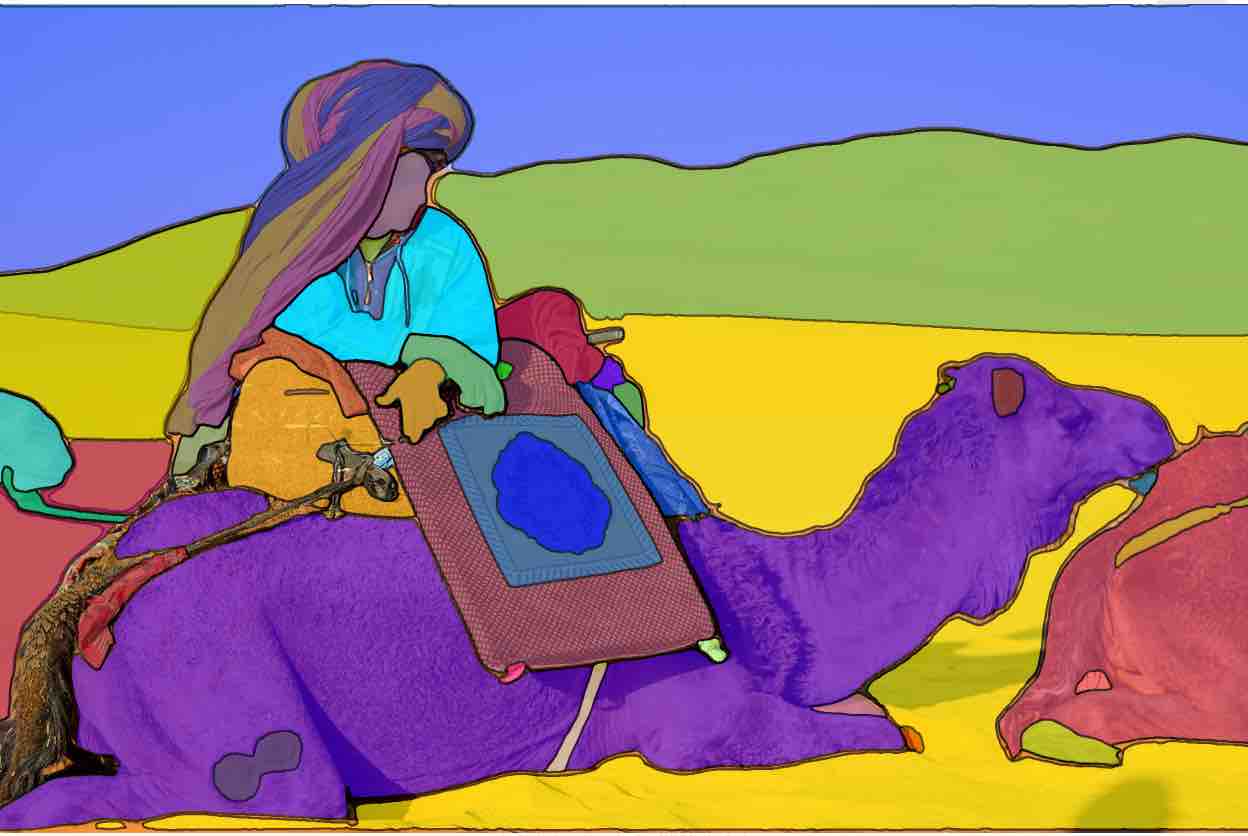}
			\centering{Annotated image}
			\label{fig:sam-dataset-sample3}
		\end{minipage}
		\begin{minipage}[c]{0.45\linewidth}
			\includegraphics[width=\linewidth]{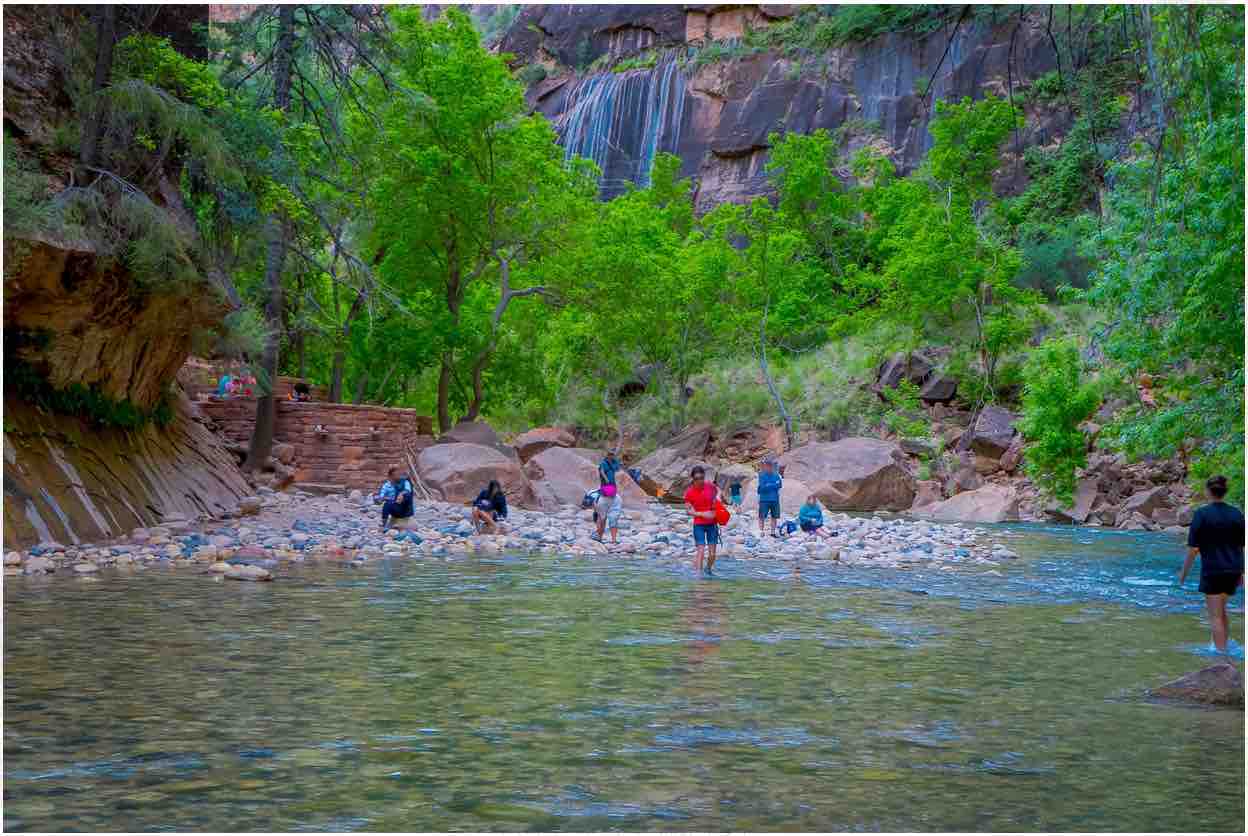}
			\centering{RGB image}
			\label{fig:sam-dataset-sample4}
		\end{minipage}
		\hfill
		\begin{minipage}[c]{0.45\linewidth}
			\includegraphics[width=\linewidth]{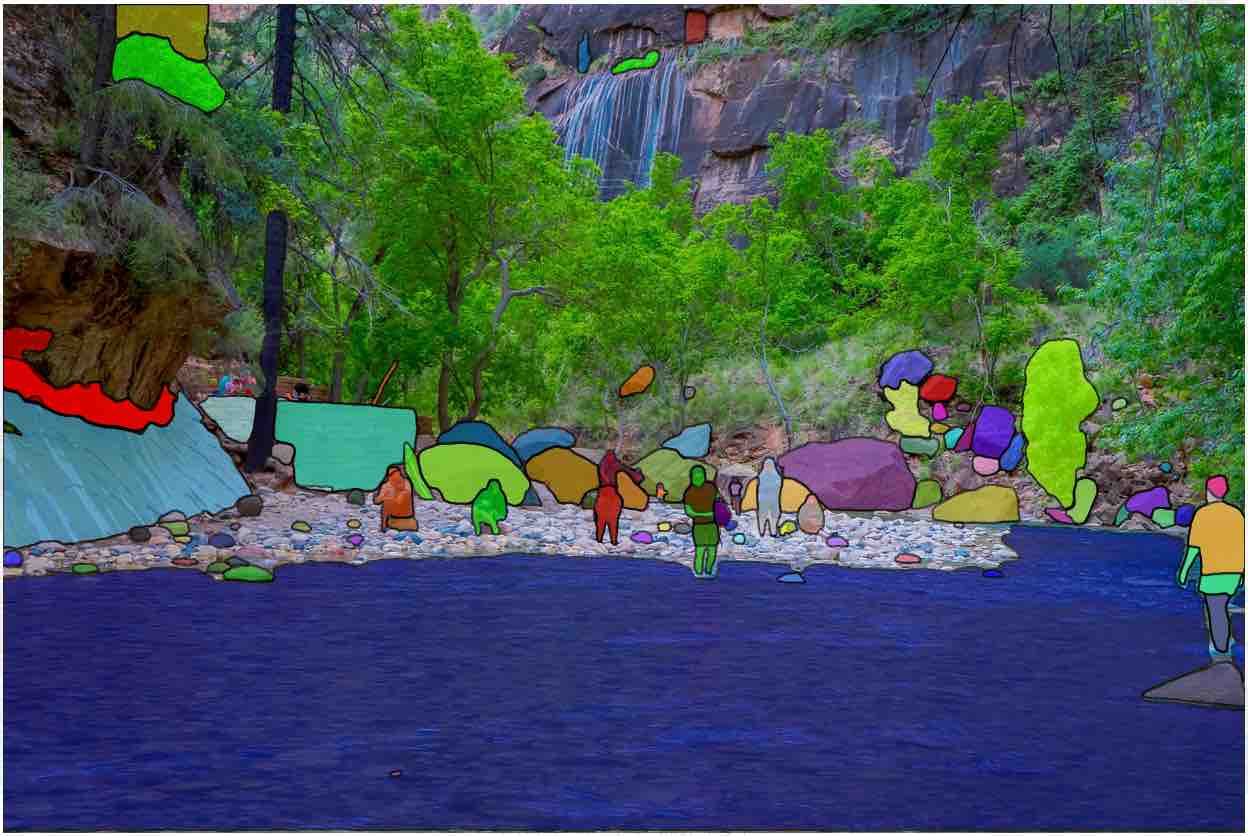}
			\centering{Annotated image}
			\label{fig:sam-dataset-sample2}
		\end{minipage}
		
		\caption{SAM dataset samples} 
		\label{fig:sam-dataset-samples}
	\end{figure}

\begin{figure}[h!]
		
		\begin{minipage}[c]{0.45\linewidth}
			\includegraphics[width=\linewidth]{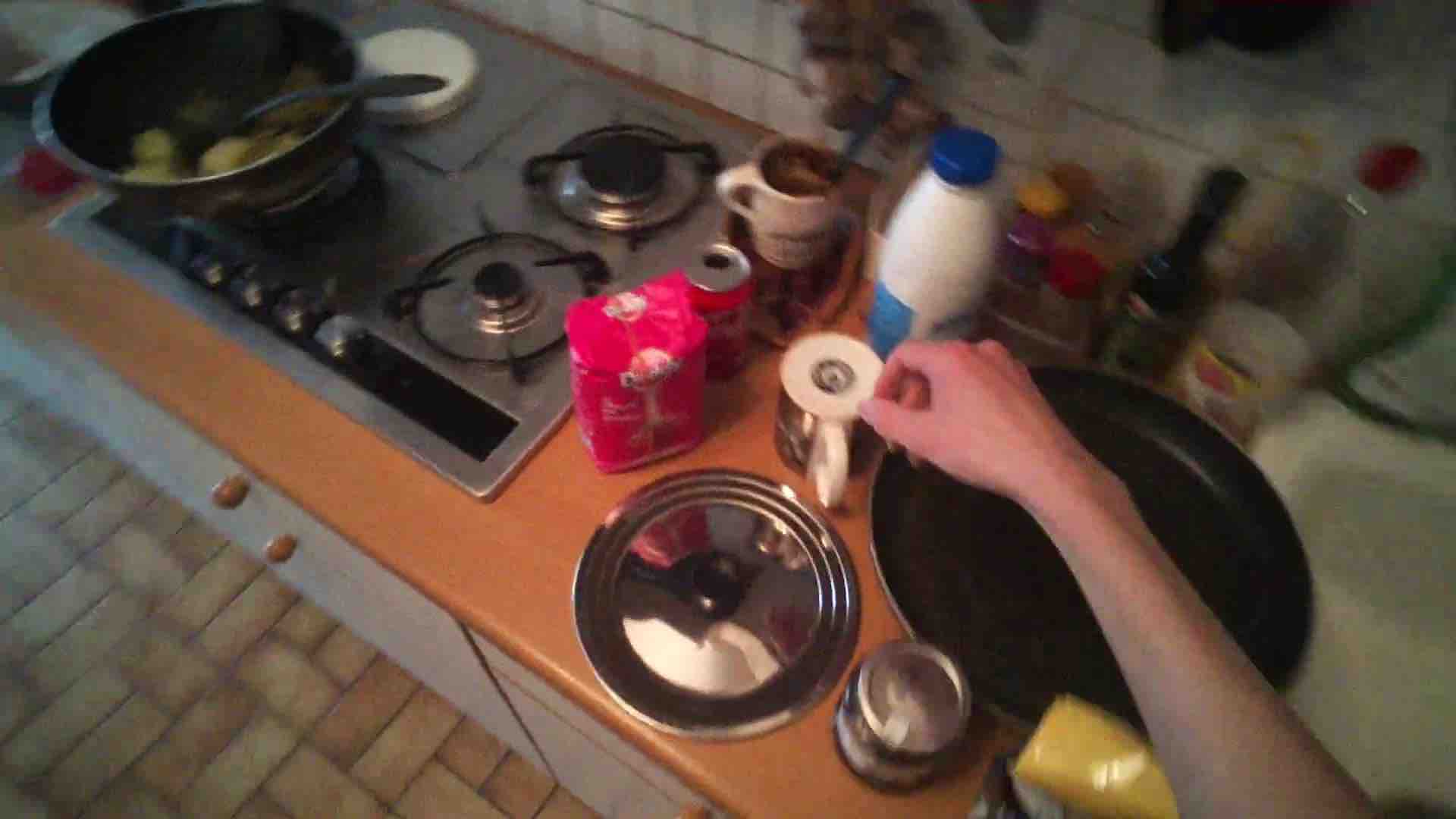}
			\centering{RGB image}
			\label{fig:gitw-dataset-sample1}
		\end{minipage}
		\hfill
		\begin{minipage}[c]{0.45\linewidth}
			\includegraphics[width=\linewidth]{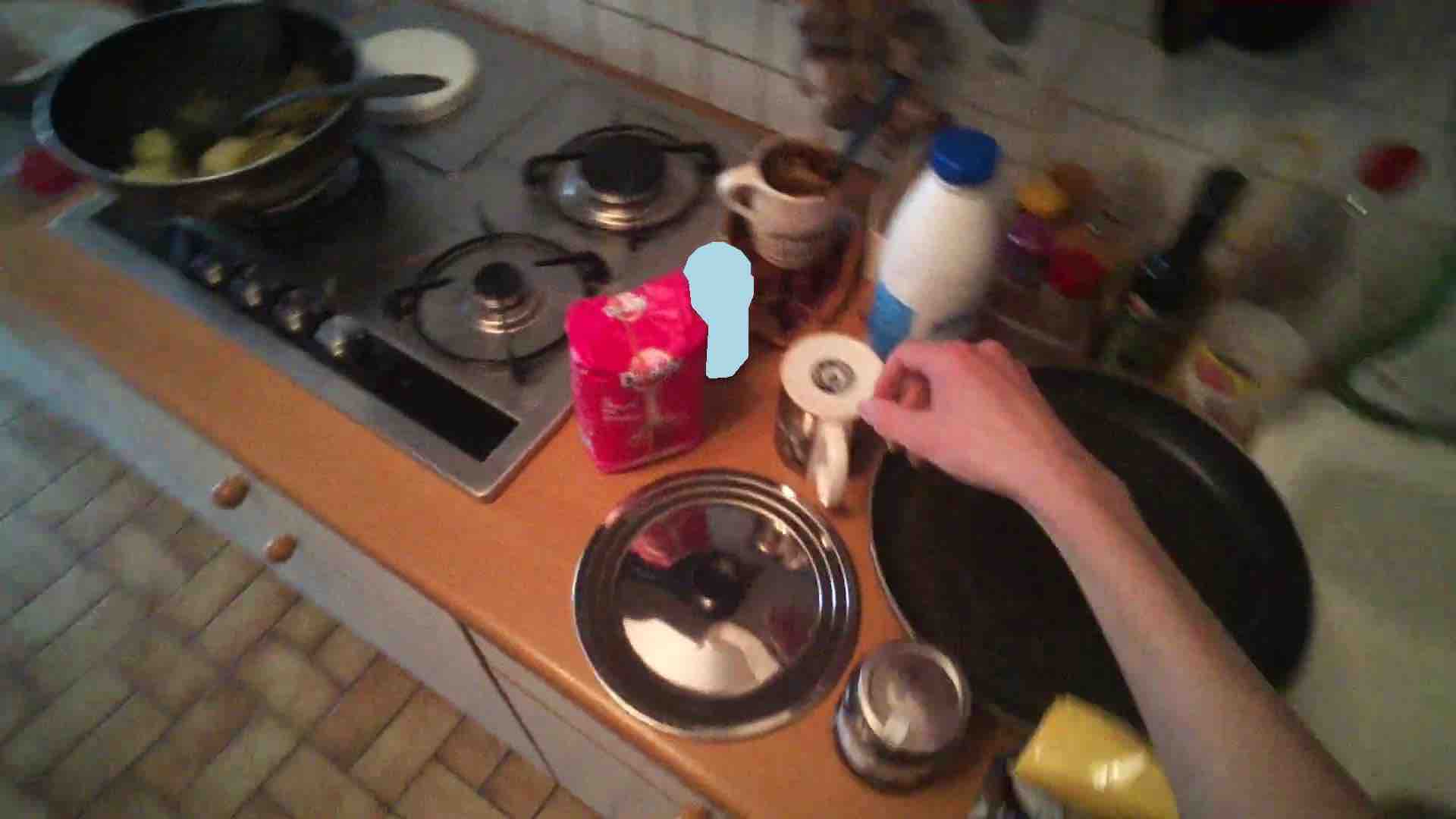}
			\centering{Annotated image (can of cola)}
			\label{fig:gitw-dataset-sample3}
		\end{minipage}
		\begin{minipage}[c]{0.45\linewidth}
			\includegraphics[width=\linewidth]{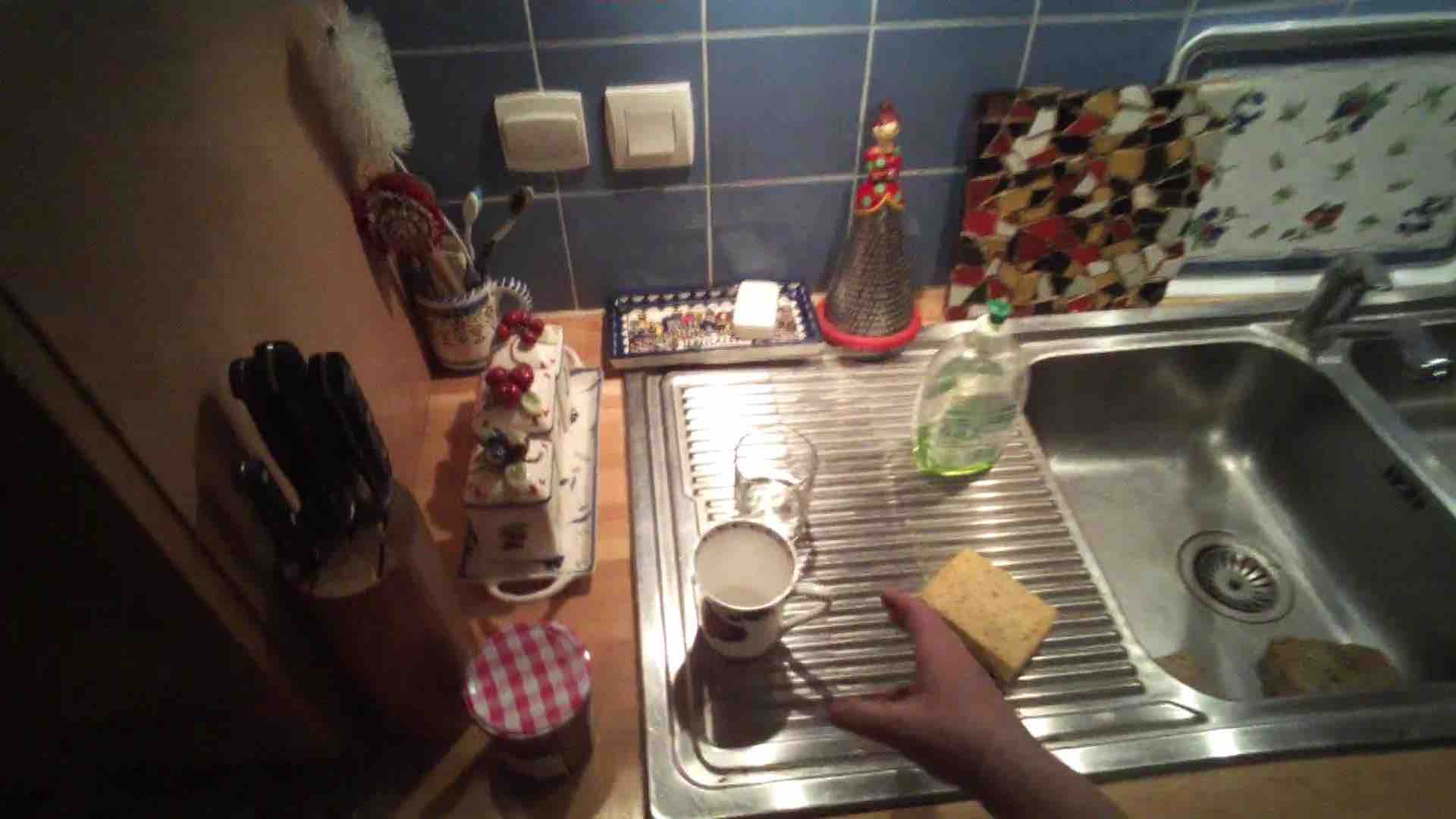}
			\centering{RGB image}
			\label{fig:gitw-dataset-sample4}
		\end{minipage}
		\hfill
		\begin{minipage}[c]{0.45\linewidth}
			\includegraphics[width=\linewidth]{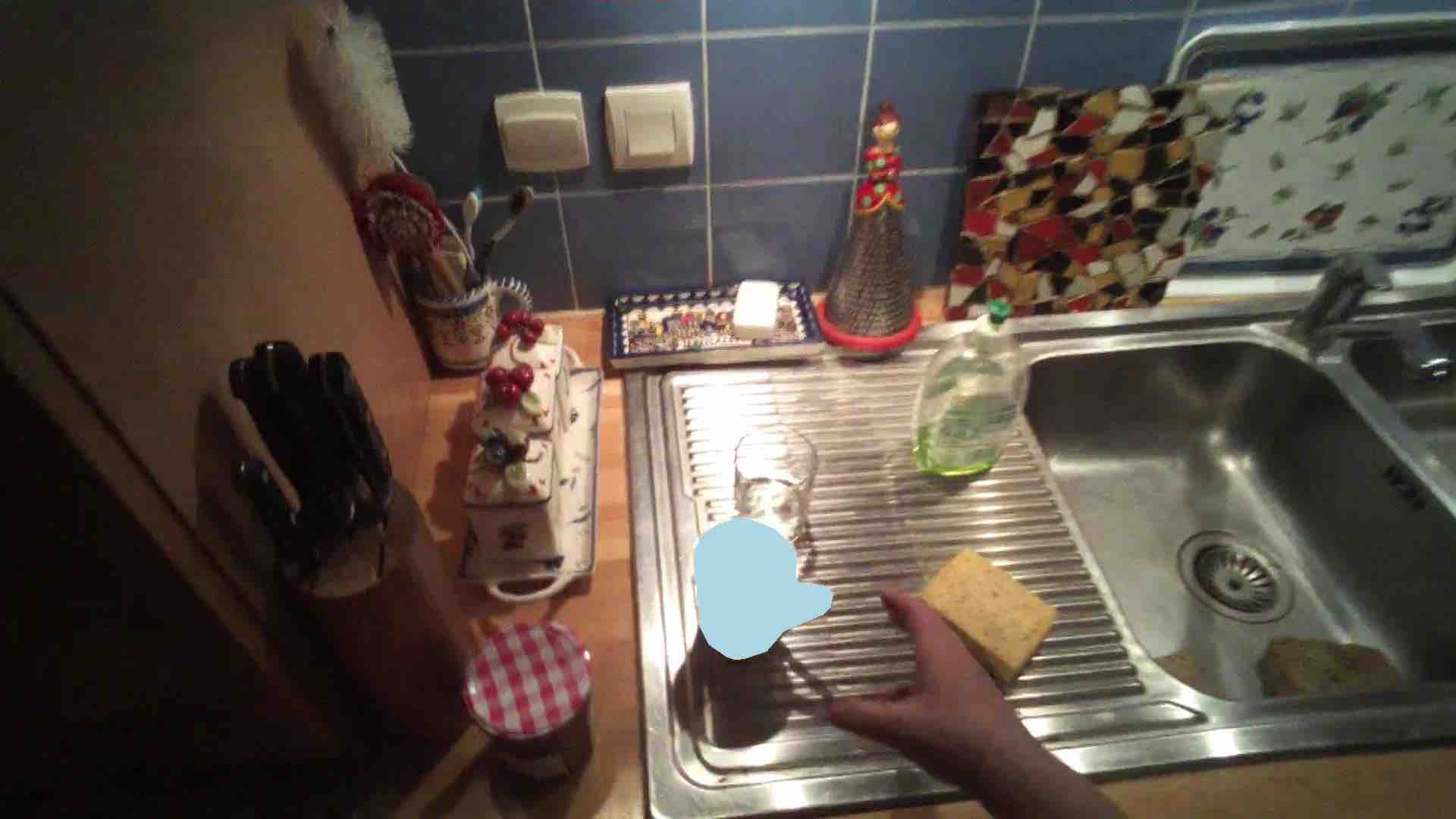}
			\centering{Annotated image (mug)}
			\label{fig:gitw-dataset-sample2}
		\end{minipage}
		
		\caption{Considered dataset samples}
		\label{fig:gitw-dataset-samples}
	\end{figure}

	Despite the appealing idea of simply applying a pre-trained model to new data, current research shows that domain adaptation is required. 
	Indeed, dataset used to train SAM  contains images of objects in low cluttered scenes and the taxonomy of SAM dataset \cite{Kirillov_2023_ICCV} is quite different from the taxonomy of objects in object grasping scenarios (see fig. \ref{fig:sam-dataset-samples} and fig. \ref{fig:gitw-dataset-samples} for comparison). Therefore, a natural step is to perform a domain adaptation. Our second approach is thus to fine-tune the pre-trained model using ground truth binary masks from the GITW dataset and apply it to the prediction of object masks. This application is modeled as:
	\begin{equation}
		\label{eqn:WithFine-Tunning}
		\text{M}(\text{D}_1) \xrightarrow{\text{fine-tune on D}_2} \text{M}_{\text{fine-tuned}}(\text{D}_{2(train)})  \rightarrow \text{D}_{2(test)}
	\end{equation}
	%$$    \text{M}(\text{D}_1) \xrightarrow{\text{fine-tune on D}_2} \text{M}_{\text{fine-tuned}}(\text{D}_{2(train)})  \rightarrow \text{D}_{2(test)} $$
	where $\text{M}(\text{D}_1)$ is the pre-trained foundation Model, $\text{D}_1$ is the Large Scale dataset, $\text{D}_{2(train)}$  is the train set of the target  dataset (GITW), $\text{M}_{fine-tuned}$ is the fine-tuned model, and $\text{D}_{2(test)}$ is the test set of the target dataset. This application is illustrated in figure \ref{fig:sam-application-finetuned}.

	\subsubsection{Fine-tuning of SAM with fuzzy masks of the GITW Dataset}
	\label{subsubsec:sam-application-fuzz-mask}

\begin{figure}[h!]
		\centering
		\includegraphics[width=0.9\textwidth]{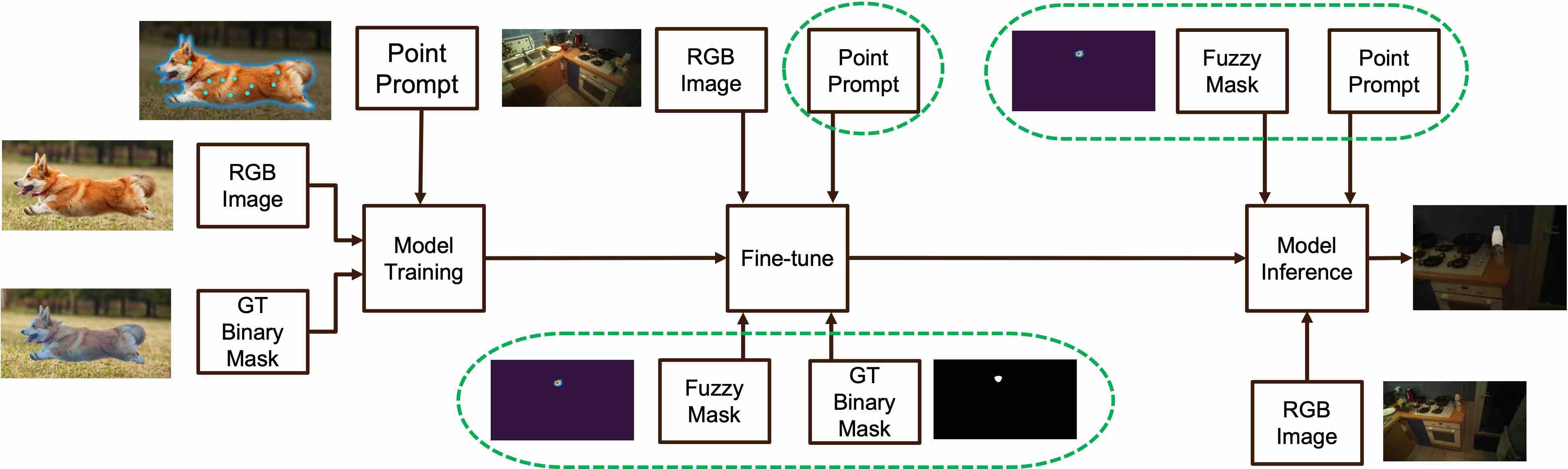}
		\caption{Application of SAM \cite{Kirillov_2023_ICCV} fine-tuned on GITW dataset}
		\label{fig:sam-application-fuzzy-mask-input}
	\end{figure}

	The third approach uses additional contextual information in the form of fuzzy masks computed in section \ref{subsec:fuzzy-mask-computation} in addition to binary masks. This is done to maximize the use of all available information to potentially improve model performance. Obtained fuzzy masks are adapted to be used as low-resolution  "mask'' inputs to the model, and fine-tuned with binary masks as ground truth. This application is implemented in  figure \ref{fig:sam-application-fuzzy-mask-input}.
	
	\subsection{Computation of ''Fuzzy'' masks}
	\label{subsec:fuzzy-mask-computation}
	For each image containing a specific object of interest to be segmented, a corresponding fuzzy mask of that object is computed at the same resolution as the image. The fuzzy mask is a probabilistic representation of the object's spatial distribution in the image. It has pixel values that range between 0 and 1, indicating the likelihood of that pixel belonging to the object, unlike a binary mask where each pixel is either fully part of the object (1) or not (0). This is done using kernel density estimation with a Gaussian kernel, so that for each image pixel, its value represents the probability density obtained from the Gaussian distribution defined by the gaze fixation points of the largest cluster, which lies on the object in the image. The "fuzziness" in this mask comes from the kernel's influence gradually decreasing with distance from the object points. Therefore, in fuzzy masks, pixels near the object points will have values close to 1, while pixels far from the points will approach zero in a smooth, tapering fashion. These probability density values are obtained using the 2D kernel density estimation:
	\begin{equation}
		\label{eqn:kde}
		f(x,y) = \frac{1}{N} \sum_{i=1}^{N} \frac{1}{2\pi h^2} \exp \left( -\frac{(x-x_i)^2 + (y-y_i)^2}{2h^2} \right)
	\end{equation}
	
	where $f(x,y)$ is the estimated probability density value at point $(x,y)$, $N$ is the number of points in the largest cluster, $(x_i, y_i)$ is  the coordinates of the $i^{th}$  point in the image and $h$ is a smoothing (spread) parameter.
	
	Due to the temporal window size used for projection, the projected and current gaze fixations only cover a limited area. Therefore, the resulting fuzzy masks do not cover the entire object area, reducing the information they contain and their use for fine-tuning the model. Therefore, a solution has been implemented to estimate an approximate bounding box around the object of interest and to sample additional points within this bounding box so that it covers the object area. The sampled points are assigned lower weights using an exponential weight decay (equation \ref{eqn:exponential-decay}) based on the euclidean distance (equation \ref{eqn:euc-distance}) between the point and the centroid of the largest cluster. 
	
	\begin{equation}
		\label{eqn:euc-distance}
		d = \sqrt{(y_i - y_c)^2 + (x_i - x_c)^2}  \text{       for i = 1, 2,..., N}
	\end{equation}
	
	\begin{equation}
		\label{eqn:exponential-decay}
		f(d) = y_0(1 - e^{\pi d})
	\end{equation}
	
	where  $d$ is the distance of the sampled point to the largest cluster centroid, $(x_i, y_i)$ is the coordinate of the sampled point, $(x_c, y_c)$ is the coordinate of the largest cluster, N is the number of sampled points, $y_0$  is the maximum weight value, $\pi$ is a decay factor, and $f(d)$ is the  assigned weight of the sampled point.

	\subsubsection{Object bounding box approximation using foveation area}
	\label{subsubsec:approx-bbox-estimation}
\begin{figure}[h!]
		
		\begin{minipage}[c]{0.45\linewidth}
			\includegraphics[width=\linewidth]{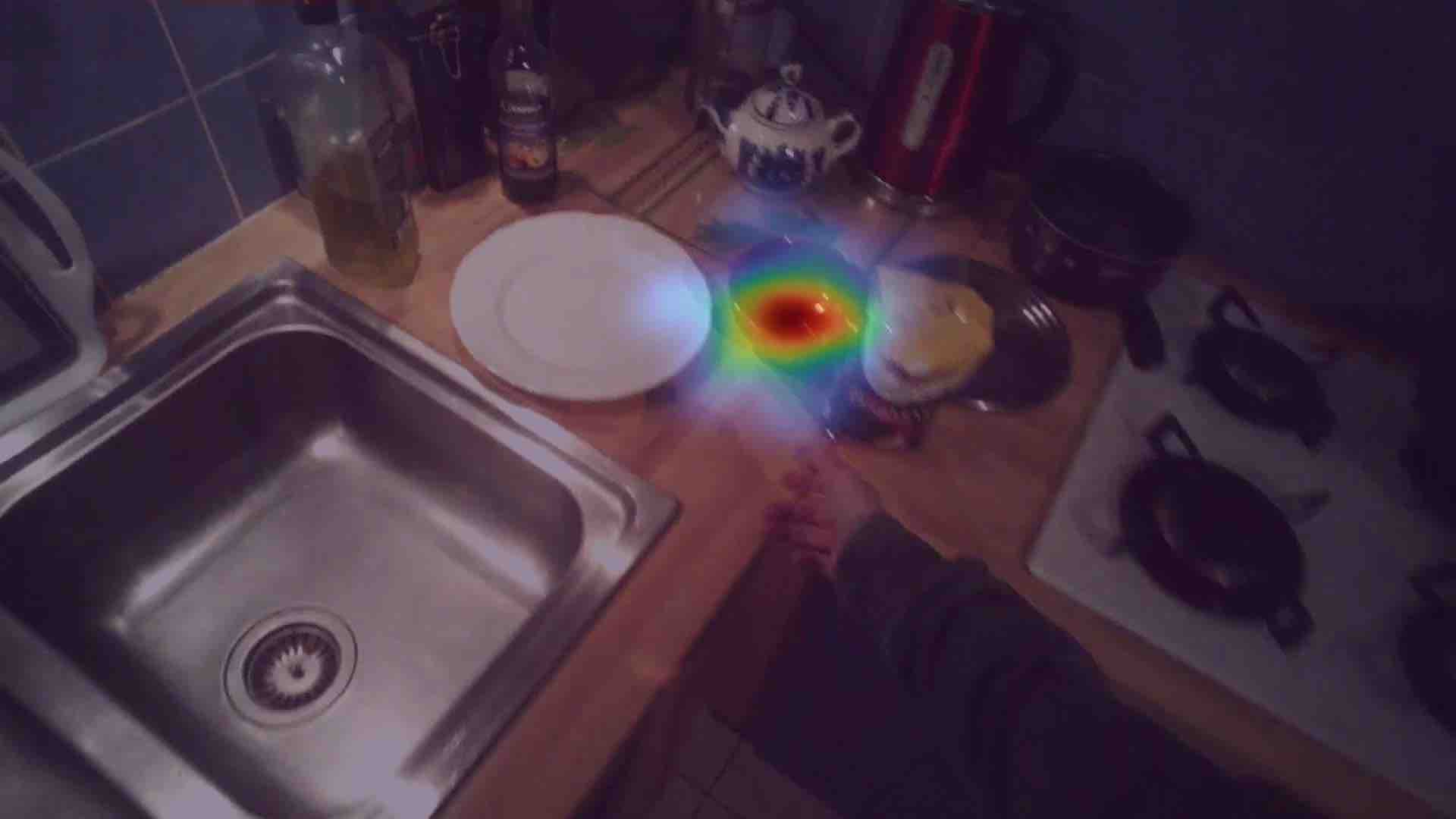}
			% \caption{}
			\label{fig:fuzzy-mask-sample1}
		\end{minipage}
		\hfill
		\begin{minipage}[c]{0.45\linewidth}
			\includegraphics[width=\linewidth]{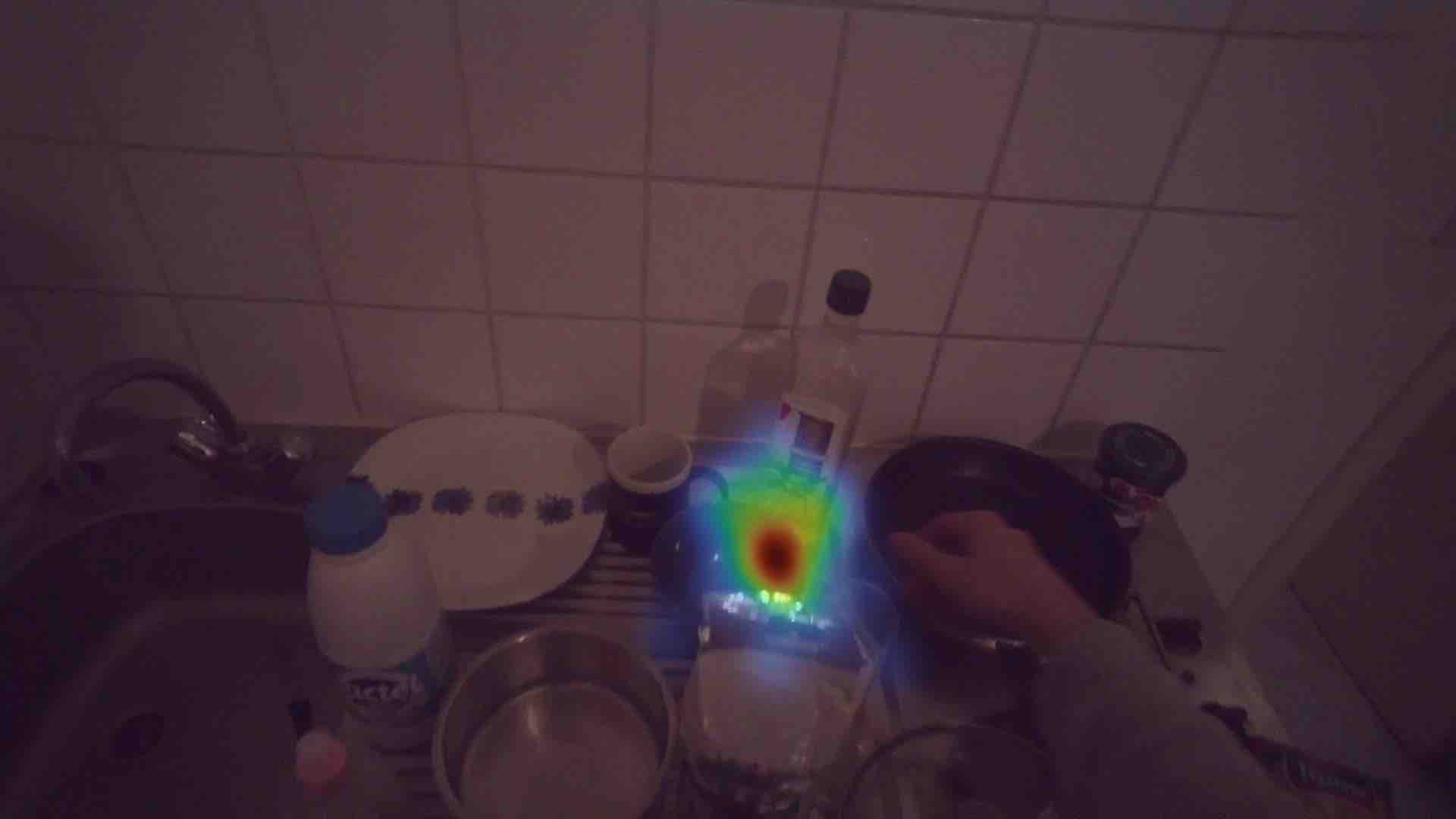}
			% \caption{}
			\label{fig:fuzzy-mask-sample2}
		\end{minipage}
		
		\begin{minipage}[c]{0.45\linewidth}
			\includegraphics[width=\linewidth]{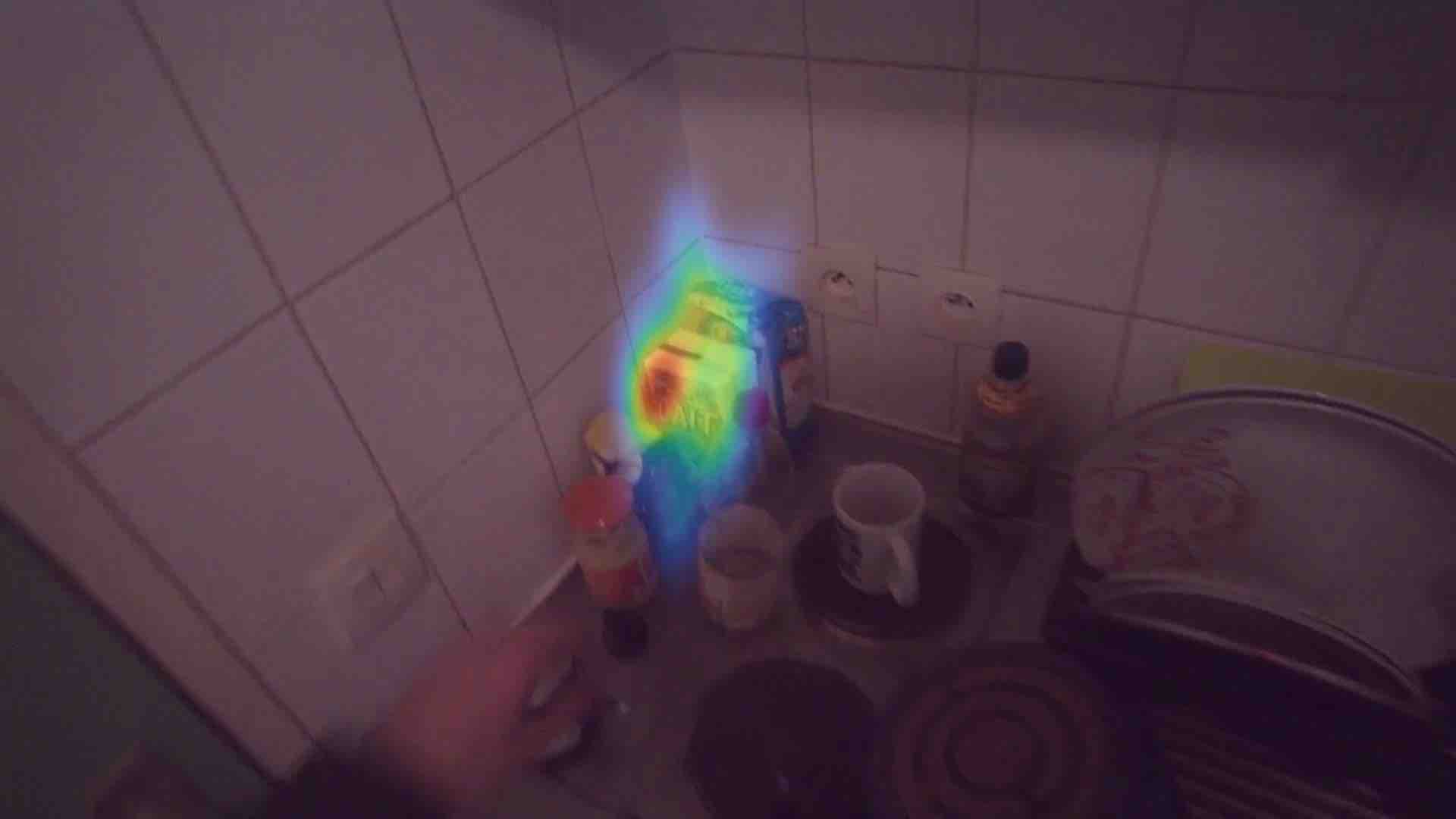}
			% \caption{}
			\label{fig:fuzzy-mask-sample3}
		\end{minipage}
		\hfill
		\begin{minipage}[c]{0.45\linewidth}
			\includegraphics[width=\linewidth]{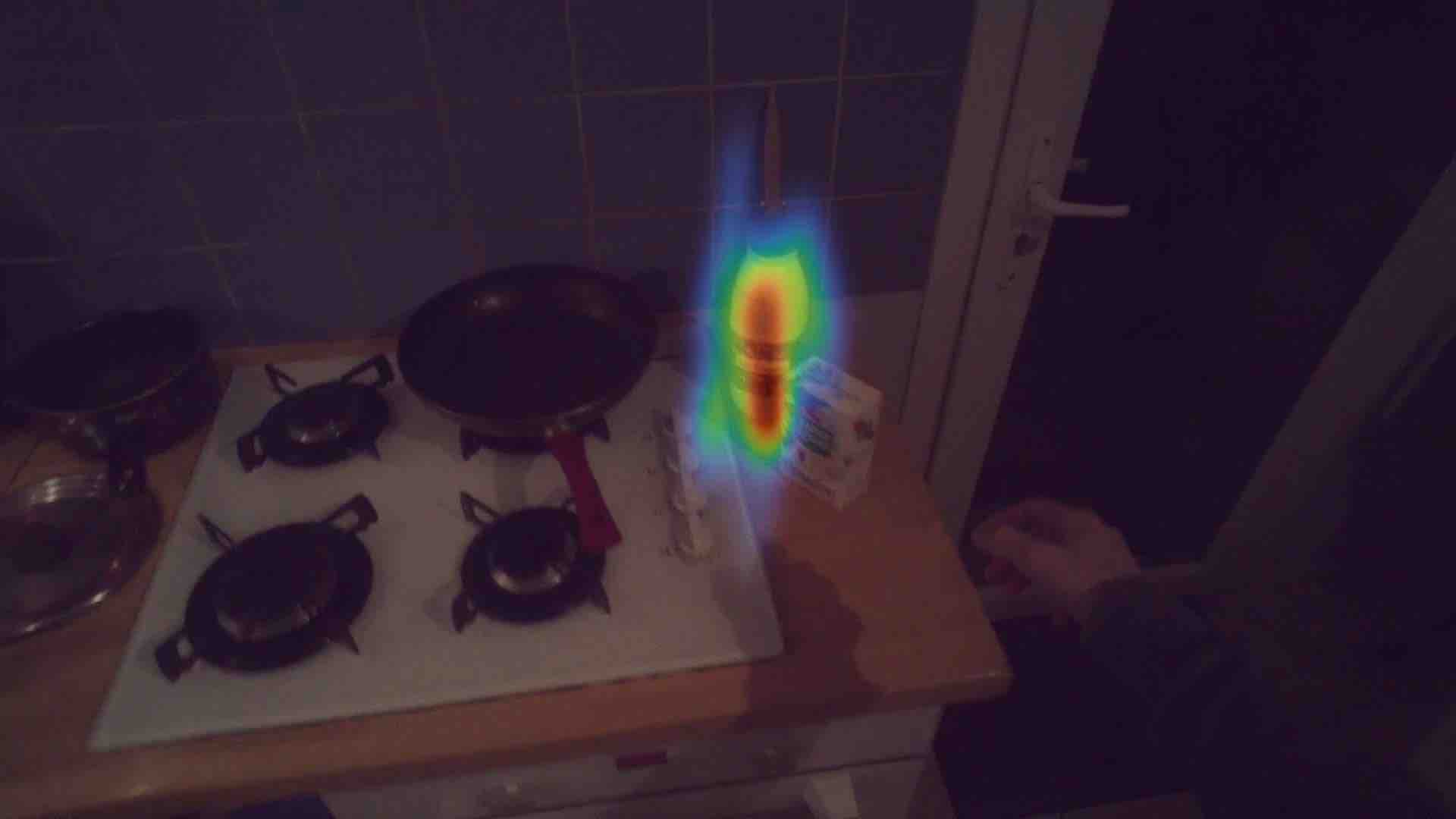}
			% \caption{}
			\label{fig:fuzzy-mask-sample4}
		\end{minipage}
		
		\caption{Computed fuzzy masks of gaze points from figure \ref{fig:projected-gaze-points}}
		\label{fig:fuzzy-mask-samples-1}
	\end{figure}

	The subject foveation area from which the bounding box is extracted is computed with the following steps:
	\begin{enumerate}
		\item \textbf{Computation of spatial extent}\\
		Using the depth of the gaze fixation point of the cluster centroid and the visual angle, the spatial extent of the gaze fixation's Gaussian distribution is computed in millimeters using equation \ref{eqn:sigma-millimeters}.
		
		\begin{equation}
			\label{eqn:sigma-millimeters}
			\sigma_{\text{mm}} = z \times \tan(\alpha)
		\end{equation}
		Where $z$ is the depth of the largest cluster's centroid, $\alpha$ - visual angle of 2 visual degrees corresponding to the fovea projection in the image plane as in \cite{ObesoBGR22}, $\sigma_{\text{mm}}$ - spatial extent of foveation (in millimeters).
		
		\item \textbf{Computation of the Field of View}\\
		The vertical and horizontal field of view in millimeters is calculated from the depth of the cluster centroid and the camera's field of view angles in both directions. This is expressed in equation \ref{eqn:fov-mm}.
		\begin{equation}
			\label{eqn:fov-mm}
			\text{FOV}_{\text{mm}} = 2z \times \tan\left(\frac{\text{FOV}_{\text{angle}}}{2}\right)
		\end{equation}
		
		where $\text{FOV}_{\text{angle}}$ is the Camera Field of View 
		
		\item \textbf{Computation of the Gaussian spatial extent:}
		The spatial extent of the Gaussian distribution on the image plane is obtained using the previously obtained spatial extent in mm $\sigma_{\text{mm}}$ and the camera's FOV (equation \ref{eqn:foveation-area-x} and equation \ref{eqn:foveation-area-y}):
		\begin{equation}
			\label{eqn:foveation-area-x}
			\sigma_{\text{px(x)}} = \sigma_{\text{mm}} \times \frac{\text{W}}{\text{FOV}_{\text{mm(x)}}}
		\end{equation}
		
		\begin{equation}
			\label{eqn:foveation-area-y}
			\sigma_{\text{px(y)}} = \sigma_{\text{mm}} \times \frac{\text{H}}{\text{FOV}_{\text{mm(y)}}}
		\end{equation}
		
		where $\sigma_{\text{px(x)}}$, $\sigma_{\text{px(y)}}$ are foveation extent in horizontal and vertical directions respectively, $W$ is the image width, and  $H$ is the image height.
		
		\item \textbf{Bounding box approximation}\\
		Using the object shape as \textit{a priori} information, a parameter $K =(K_x, K_y)$ is set in both directions to approximate the object bounding box based on the subject foveation area (eq. \ref{eqn:bbox-approximation-x} and eq. \ref{eqn:bbox-approximation-y} ).

		\begin{equation}
			\label{eqn:bbox-approximation-x}
			x_{\text{{min}}}, x_{\text{max}} = x_{c} \pm K_x \times \sigma_\text{px(x)}
		\end{equation}
		
		\begin{equation}
			\label{eqn:bbox-approximation-y}
			y_{\text{{min}}}, y_{\text{max}} = y_{c} \pm K_y \times \sigma_\text{px(y)}
		\end{equation}
		where, $(x_{min}, y_{min})$  is the top-left corner of bounding box, $(x_{max}, y_{max})$ is the bottom-right corner of bounding box, $(x_c, y_c)$ - coordinates of centroid of the largest cluster,  and ($K_x, K_y$) is the object shape parameter.
	\end{enumerate}
	These built fuzzy masks are adapted to the shapes of the objects (see fig. \ref{fig:fuzzy-mask-samples-1}) and may serve as additional prompt for the SAM model.

	\section{Results and Analysis}
	We present in this section the  dataset "Grasping-In-The-Wild"
	(\texttt{GITW}), that will be used in the presented experiments on object segmentation in cluttered scenes. We also detail the parameterization of our methods, and the analysis of the results.  
	\label{sec:results}
	\subsection{The  \texttt{GITW} dataset} 
	\label{subsec:dataset}
	The  \texttt{GITW} dataset corresponds to our object grasping scenario. It is publicly available on the NAKALA server \cite{graspingInTheWild}. It was captured using Tobii Glasses 2 \cite{tobiiProGlasses2} during an experiment in which five subjects located and grasped objects in seven natural kitchen environments. The object taxonomy contains sixteen different classes, and for each instance of the object classes it contains a video file, a \textit{json} file with gaze fixations, and foreground and random background patches. 
	Note that for our purpose of precisely evaluating segmentation quality, we could not use this latter information and instead had to annotate object masks with pixel precision.
	We have only considered frames before the grasping time, as required by our object segmentation scenario. The overall number of object instances is depicted in table \ref{tab:frames-breakdown}. It is important to note that the object classes exhibit strong variability in appearance.

	\begin{table}[h!]
		
		\resizebox{1\width}{!}{
			\begin{tabular}{lcc}
				\hline
				& \textbf{Object Class}                     & \textbf{Considered frames} \\ \hline
				1                            & Bowl                                      & 522                        \\
				% \rowcolor[HTML]{C0C0C0} 
				2                            & Can of Coca Cola                          & 646                        \\
				3                            & Frying Pan                                & 897                        \\
				% \rowcolor[HTML]{C0C0C0} 
				4                            & Glass                                     & 767                        \\
				5                            & Jam                                       & 517                        \\
				% \rowcolor[HTML]{C0C0C0} 
				6                            & Lid                                       & 1108                       \\
				7                            & Milk Bottle                               & 534                        \\
				% \rowcolor[HTML]{C0C0C0} 
				8                            & Mug                                       & 1318                       \\
				9                            & Oil Bottle                                & 736                        \\
				% \rowcolor[HTML]{C0C0C0} 
				10                           & Plate                                     & 1223                       \\
				11                           & Rice                                      & 565                        \\
				% \rowcolor[HTML]{C0C0C0} 
				12                           & Saucepan                                  & 663                        \\
				13                           & Sponge                                    & 1063                       \\
				% \rowcolor[HTML]{C0C0C0} 
				14                           & Sugar                                     & 1000                       \\
				15                           & Vinegar Bottle                            & 1121                       \\
				% \rowcolor[HTML]{C0C0C0} 
				16                           & Wash Liquid                               & 706                        \\ \hline
				\textbf{Total}               & \textbf{}                                & \textbf{13345}             \\ \hline
			\end{tabular}%
		}
		\caption{Number of Frames considered for each class in object taxonomy}
		\label{tab:frames-breakdown}
		
	\end{table}

	For pixel-wise annotation of object masks,  we designed a large-scale campaign for the annotation of sixteen object classes on the web-based, collaboration-friendly platform Roboflow \footnote{https://universe.roboflow.com}. 
	The Roboflow graphical interface allows annotators to click on the object of interest in each frame, and proposes a semantic segmentation of the selected object. Roboflow provides an interactive refinement by the user of the proposed area. It utilizes the Segment Anything model under the hood, where the selected image points serve as foreground or background points to the model for segmentation. The annotated dataset can be found at \footnote{https://universe.roboflow.com/iwrist/grasping-in-the-wild}. 
	Regarding the selection of foreground points as essential prompts for the concerned model, we adopted our proposed approach, see section \ref{subsec:gaze-points}. In the SAM fine-tuning experiment, we shuffled the samples for each object class and split them into 70\% for training and 30\% for testing. 
	
	% \subsection{Dataset Annotation}
	% \label{subsec:dataset-annotation}
	% The successful completion of the foundation model adaptation to segmenting objects in the wild requires a means of evaluating the model's performance on the dataset. This involves comparing the predicted masks to ground-truth masks using the specified metrics. The dataset as at the commencement of the experiment did not contain Ground-truth mask information, hence annotation of the dataset was performed to obtain Ground-truth binary masks of the objects in the dataset. 

	\subsection{Evaluation metrics}
	\label{subsec:eval-metrics}
	To evaluate the precision of extracted segmentation masks, the most popular metrics are: intersection over union (IoU), pixel accuracy and DICE coefficient \cite{Muller2022}. In our work, we use only the IoU metric. Pixel accuracy measures the ratio of correctly classified pixels over the number of all image pixels. However, due to our small object sizes relative to frame resolution, pixel accuracy suffers from class imbalance. The DICE coefficient metric is similar to IoU and is mainly used in medical image segmentation. 
	
	The IoU metric measures the overlap between the predicted and ground-truth masks. \
	For a video frame $f$, it is given by:
	\begin{equation}
		IoU_f = \frac{{\mid M \cap \tilde{M} \mid}}{{\mid M \cup \tilde{M} \mid}}
	\end{equation}
	Where $\tilde{M}$ is the predicted binary mask, $M$ is the ground-truth binary mask, $\mid M \cap \tilde{M} \mid$ is the number of pixels in the masks' intersection area, $\mid M \cup \tilde{M} \mid$ is the number of pixels in the masks' union area.
	The mean IoU for the considered grasping sequence of $N$ frames $f_i$ is given by:
	\begin{equation}
		mIoU = \frac{1}{N} \sum_{i=1}^{N} IoU_{f_i}
	\end{equation}

	\subsection{Experiment setup}
	\label{subsec:exp-config}
	The setup of the experiment conducted is provided below.
	
	\begin{itemize}
		\item \textbf{SAM Architecture}\\
		The chosen bakbone architecture was the ViT-Base model \cite{Kirillov_2023_ICCV}. Here the ViT-Base model is employed as image encoder. It is the lightest of the three models  available for SAM : ViT-Base, ViT-Large, and ViT-Heavy and contains 91M parameters.
		
		\item \textbf{Software and Hardware Specifications}\\
		The experiment was performed using a Tesla P100 GPU with 16GB of graphics memory. The fine-tuning process involved freezing the parameters of both encoder models--image and prompt-- so that only the weights of the mask decoder were updated.  
		
		\item \textbf{Experiment Hyperparameters}\\
		The following values were set for hyperparameters: \\fixed Learning rate: $1e^{-5}$, Optimizer: Adam, Number of epochs: 30, Batch size: 4 and Number of object points: 5.
	\end{itemize}
	
	\subsection{Quantitative Evaluation}
	\label{subsec:quantitative-eval}
	This section utilizes the evaluation metric IoU from section \ref{subsec:eval-metrics} to quantify the performance of the proposed approaches, on the GITW dataset.
	
	\subsubsection{Evaluation of pre-trained and binary mask fine-tuned models}
	\label{subsubsec:pre-trained-vs-bin-mask-models}
	
\begin{figure}[h!]
		\centering
		\includegraphics[width=0.8\linewidth,keepaspectratio]{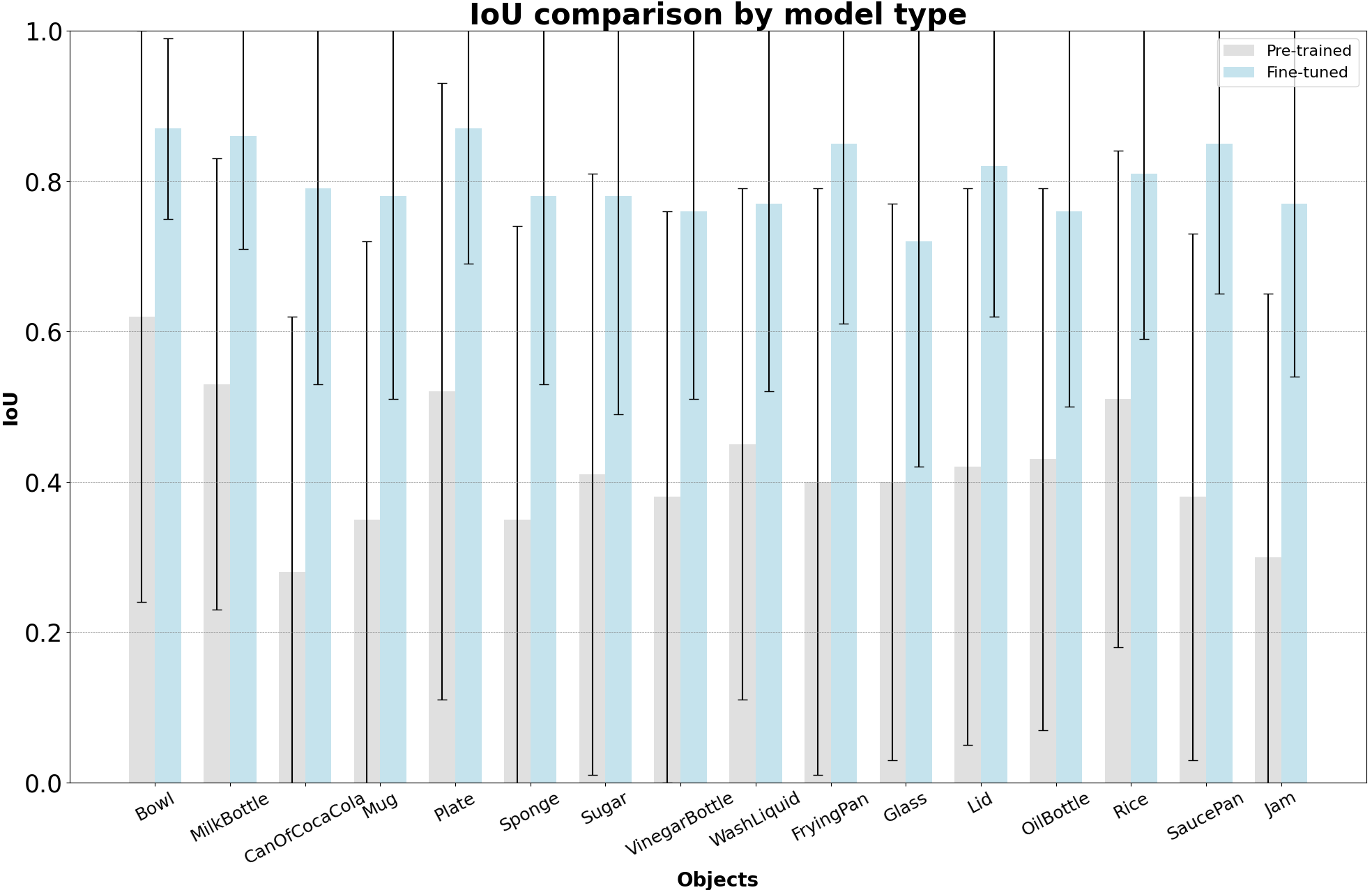}
		\caption{IoU evaluation comparison of binary masks for pre-trained and fine-tuned models (blue bars for the fine-tuned model and grey bars for the pre-trained one)}
		\label{fig:res-pre-trained-bin-mask}
	\end{figure}

	The evaluation of the fine-tuned model with ground-truth binary masks is done by comparing it with the pre-trained model. This is illustrated in figure \ref{fig:res-pre-trained-bin-mask}, and detailed in table \ref{tab:eval-results-bin-mask-fine-tuned}. For the pre-trained model, the mean \textit{IoU} values range from 0.28 (Can of Coca Cola) to 0.62 (Bowl), with a standard deviation for \textit{IoU} being highest for the ``Sugar'' object at 0.40 and lowest for the ``Milk Bottle'' object at 0.30, indicating different levels of segmentation variability. In contrast, the fine-tuned model shows quite a strong increase in performance. The mean \textit{IoU} values now range from 0.72 (Glass) to 0.87 (Bowl and Plate). The standard deviation for \textit{IoU} decreases across the board, with the highest being 0.30 for ``Glass'' and the lowest at 0.12 for ``Bowl'', suggesting more reliable segmentation results.
	
	These improvements are particularly noticeable for the ``Can of Coca Cola'', where the \textit{IoU} increases from 0.28 to 0.79. Similarly, for the ``Mug'' the \textit{IoU} increases from 0.35 to 0.78. The reduced standard deviations indicate better stability and reliability in the model's performance across all objects. The overall performance shows that the fine-tuning process improves segmentation accuracy and consistency, particularly for objects that were previously difficult to segment, such as the ``Can of Coca Cola'' and ``Sponge''.  It can be seen that when fine-tuned to the target corpus of natural objects in highly cluttered scenes, the model's performance improves across all object classes considered, up to 0.51 points for the ``Can of Coca Cola'' object. To push the investigation to its limits, we also evaluated the model without any prompts, and the resulting $mIoU$ is 0.0. 
	
	\begin{table}[h!]
		% \footnotesize
		\resizebox{\width}{!}{%
			% \footnotesize
			\begin{tabular}{lccc}
				\hline
				& \multicolumn{3}{c}{\textbf{mIoU}}                                                                                                          \\ \cline{2-4} 
				\multirow{-2}{*}{\textbf{}} & \textbf{Pre-trained Model}                   &                                              & \textbf{Fine-tuned with binary mask}         \\ \hline
				Bowl                        & $0.62 \pm 0.38$                              &                                              & $\mathbf{0.87 \pm 0.12}$                     \\
				% \rowcolor[HTML]{C0C0C0} 
				Can of Coca Cola            & $0.28 \pm 0.34$                              &                                              & $\mathbf{0.79 \pm 0.26}$                     \\
				Frying Pan                  & $0.40 \pm 0.39$                              &                                              & $\mathbf{0.85 \pm 0.24}$                     \\
				% \rowcolor[HTML]{C0C0C0} 
				Glass                       & $0.40 \pm 0.37$                              &                                              & $\mathbf{0.72 \pm 0.30}$                     \\
				Jam                         & $0.30 \pm 0.35$                              &                                              & $\mathbf{0.77 \pm 0.23}$                     \\
				% \rowcolor[HTML]{C0C0C0} 
				Lid                         & $0.42 \pm 0.37$                              &                                              & $\mathbf{0.82 \pm 0.20}$                     \\
				Milk Bottle                 & $0.53 \pm 0.30$                              &                                              & $\mathbf{0.86 \pm 0.15}$                     \\
				% \rowcolor[HTML]{C0C0C0} 
				Mug                         & $0.35 \pm 0.37$                              &                                              & $\mathbf{0.78 \pm 0.27}$                     \\
				Oil Bottle                  & $0.43 \pm 0.36$                              &                                              & $\mathbf{0.76 \pm 0.26}$                     \\
				% \rowcolor[HTML]{C0C0C0} 
				Plate                       & $0.52 \pm 0.41$                              &                                              & $\mathbf{0.87 \pm 0.18}$                     \\
				Rice                        & $0.51 \pm 0.33$                              &                                              & $\mathbf{0.81 \pm 0.22}$                     \\
				% \rowcolor[HTML]{C0C0C0} 
				Saucepan                    & $0.38 \pm 0.35$                              &                                              & $\mathbf{0.85 \pm 0.20}$                     \\
				Sponge                      & $0.35 \pm 0.39$                              &                                              & $\mathbf{0.78 \pm 0.25}$                     \\
				% \rowcolor[HTML]{C0C0C0} 
				Sugar                       & $0.41 \pm 0.40$                              &                                              & $\mathbf{0.78 \pm 0.29}$                     \\
				Vinegar Bottle              & $0.38 \pm 0.38$                              &                                              & $\mathbf{0.76 \pm 0.25}$                     \\
				% \rowcolor[HTML]{C0C0C0} 
				Wash Liquid                 & $0.45 \pm 0.34$                              &                                              & $\mathbf{0.77 \pm 0.25}$                     \\ \hline
			\end{tabular}%
			
		}
		\caption{Performance of binary mask fine-tuned model against pre-trained model on GITW Dataset}
		\label{tab:eval-results-bin-mask-fine-tuned}
	\end{table}
	
	\subsubsection{Evaluation of fine-tuned model with  supplementary fuzzy-masks}
	\label{subsubsec:fuzzy-mask-input-eval}
\begin{figure}[h!]
		\centering
		\includegraphics[width=0.75\linewidth,keepaspectratio]{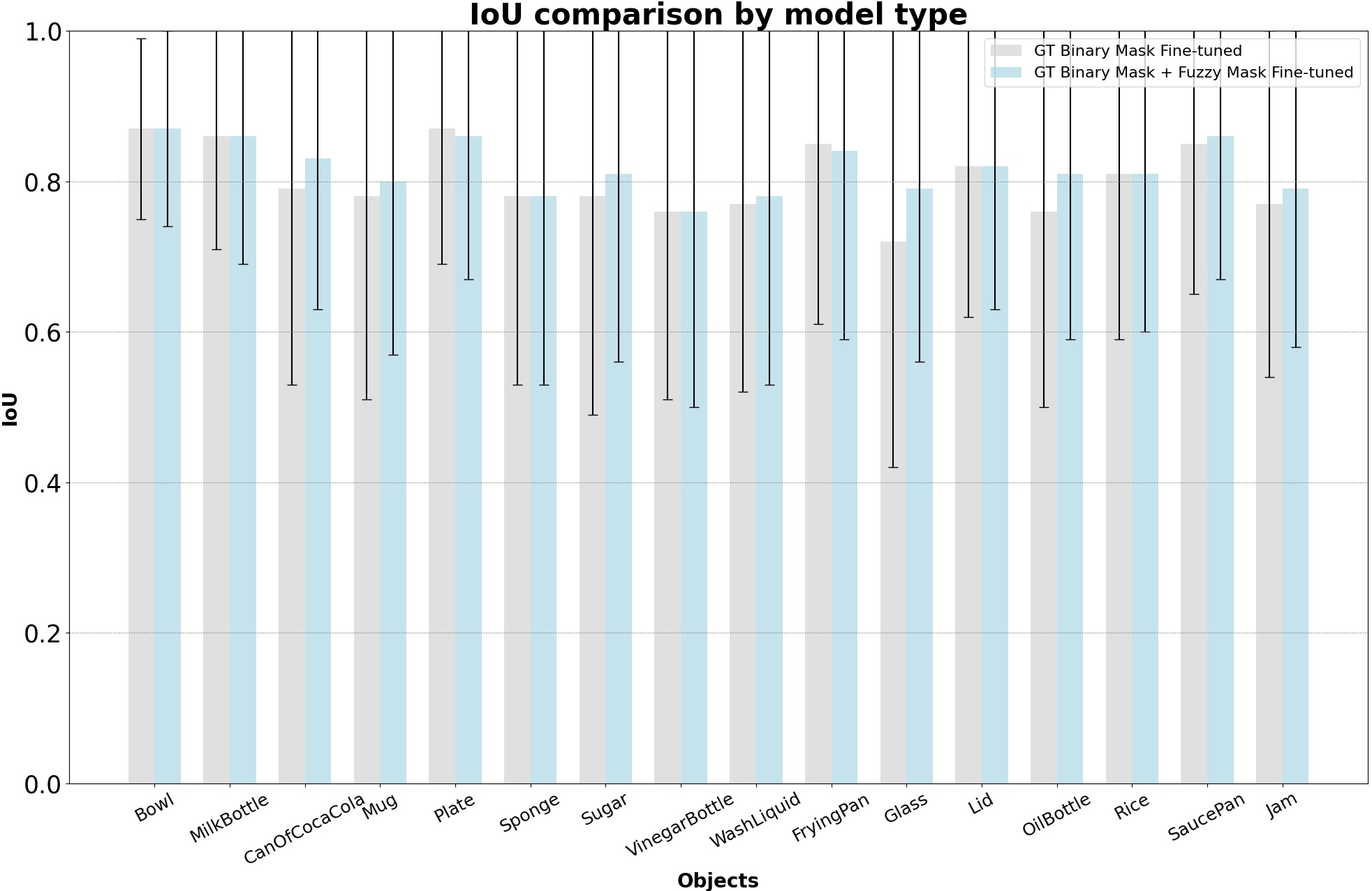}
		\caption{IoU evaluation comparison of both fine-tuned models}
		\label{fig:res-bin-mask-fuzzy-mask}
	\end{figure}

	The fine-tuned model using the binary masks as ground-truth and the fuzzy masks as additional input is evaluated against the fine-tuned model using only the binary masks, see section \ref{subsubsec:pre-trained-vs-bin-mask-models}. The result of this fine-tuning approach, shown in figure \ref{fig:res-bin-mask-fuzzy-mask}, shows a slight improvement in performance for the mean IoU metric, with ``Can of Coca Cola'' increasing from 0.79 to 0.83 and ``Oil Bottle'' increasing from 0.76 to 0.81. The stability of the predictions also increased, with the upper bound of the standard deviation decreasing from 0.30 to 0.26. This suggests a more stable and reliable performance across different objects.
	
	Additionally, both models are compared in terms of trade-off between increased performance and effects on inference computation time. These results are presented in table \ref{tab:time-comparison}, and show that with the computation and inclusion of the fuzzy mask as additional input prompt to the segmentation model, a slight increase in IoU for 8/16 objects considered is observed. This however comes with substantial increase in the computation time at the inference stage. Due to the time constraints surrounding the application area of the model, the target is to be as close to real time as possible, therefore, for the rest of the paper, we present experiments without the use of fuzzy masks as additional input prompts during segmentation.
	
	\begin{table}[h!]
		\centering
		\resizebox{\columnwidth}{!}{%
			\begin{tabular}{lllllll}
				\hline
				\multicolumn{1}{c}{\textbf{}} & \multicolumn{3}{c}{\textbf{Time (s)}}                                                                                                & \multicolumn{3}{c}{\textbf{mIoU}}                                                                                                    \\ \cline{2-7} 
				\multicolumn{1}{c}{\textbf{}} & \multicolumn{1}{c}{\textbf{No Fuzzy Mask Input}} & \multicolumn{1}{c}{\textbf{Fuzzy Mask Input}} & \multicolumn{1}{c}{\textbf{\% $\Delta$}} & \multicolumn{1}{c}{\textbf{No Fuzzy Mask Input}} & \multicolumn{1}{c}{\textbf{Fuzzy Mask Input}} & \multicolumn{1}{c}{\textbf{\% $\Delta$}} \\ \hline
				Bowl                          & 0.43 $\pm$ 0.05                                     & \textbf{4.01 $\pm$ 0.25}                         & +832.56                           & \textbf{0.87 $\pm$ 0.12}                            & 0.87 $\pm$ 0.13                                  & 0.00                              \\
				Can of Coca Cola              & 0.42 $\pm$ 0.06                                     & \textbf{4.35 $\pm$ 0.12}                         & +935.71                           & 0.79 $\pm$ 0.26                                     & \textbf{0.83 $\pm$ 0.20}                         & +5.06                             \\
				Frying Pan                    & 0.43 $\pm$ 0.05                                     & \textbf{3.36 $\pm$ 0.15}                         & +681.40                           & \textbf{0.85 $\pm$ 0.24}                            & 0.84 $\pm$ 0.25                                  & -1.18                             \\
				Glass                         & 0.43 $\pm$ 0.05                                     & \textbf{4.51 $\pm$ 0.16}                         & +948.84                           & 0.72 $\pm$ 0.30                                     & \textbf{0.79 $\pm$ 0.23}                         & +9.72                             \\
				Jam                           & 0.43 $\pm$ 0.05                                     & \textbf{4.47 $\pm$ 0.18}                         & +939.53                           & 0.77 $\pm$ 0.23                                     & \textbf{0.79 $\pm$ 0.21}                         & +2.60                             \\
				Lid                           & 0.43 $\pm$ 0.07                                     & \textbf{4.53 $\pm$ 0.22}                         & +953.49                           & 0.82 $\pm$ 0.20                                     & \textbf{0.82 $\pm$ 0.19}                         & 0.00                              \\
				Milk Bottle                   & 0.42 $\pm$ 0.06                                     & \textbf{4.48 $\pm$ 0.23}                         & +966.67                           & \textbf{0.86 $\pm$ 0.15}                            & 0.86 $\pm$ 0.17                                  & 0.00                              \\
				Mug                           & 0.42 $\pm$ 0.06                                     & \textbf{4.55 $\pm$ 0.13}                         & +983.33                           & 0.78 $\pm$ 0.27                                     & \textbf{0.8 $\pm$ 0.23}                          & +2.56                             \\
				Oil Bottle                    & 0.42 $\pm$ 0.06                                     & \textbf{4.26 $\pm$ 0.17}                         & +914.29                           & 0.76 $\pm$ 0.26                                     & \textbf{0.81 $\pm$ 0.22}                         & +6.58                             \\
				Plate                         & 0.43 $\pm$ 0.05                                     & \textbf{3.81 $\pm$ 0.29}                         & +786.05                           & \textbf{0.87 $\pm$ 0.18}                            & 0.86 $\pm$ 0.19                                  & -1.15                             \\
				Rice                          & 0.43 $\pm$ 0.05                                     & \textbf{4.46 $\pm$ 0.15}                         & +937.21                           & 0.81 $\pm$ 0.22                                     & \textbf{0.81 $\pm$ 0.21}                         & 0.00                              \\
				Saucepan                      & 0.42 $\pm$ 0.05                                     & \textbf{3.92 $\pm$ 0.29}                         & +833.33                           & 0.85 $\pm$ 0.20                                     & \textbf{0.86 $\pm$ 0.19}                         & +1.18                             \\
				Sponge                        & 0.43 $\pm$ 0.06                                     & \textbf{4.54 $\pm$ 0.12}                         & +955.81                           & 0.78 $\pm$ 0.25                                     & 0.78 $\pm$ 0.25                                  & 0.00                              \\
				Sugar                         & 0.42 $\pm$ 0.05                                     & \textbf{4.15 $\pm$ 0.26}                         & +888.10                           & 0.78 $\pm$ 0.29                                     & \textbf{0.81 $\pm$ 0.25}                         & +3.85                             \\
				Vinegar Bottle                & 0.42 $\pm$ 0.06                                     & \textbf{4.31 $\pm$ 0.17}                         & +926.19                           & \textbf{0.76 $\pm$ 0.25}                            & 0.76 $\pm$ 0.26                                  & 0.00                              \\
				Wash Liquid                   & 0.42 $\pm$ 0.06                                     & \textbf{4.41 $\pm$ 0.12}                         & +950.00                           & 0.77 $\pm$ 0.25                                     & \textbf{0.78 $\pm$ 0.25}                         & +1.30                             \\ \hline
			\end{tabular}%
		}
		\caption{Comparison of Inference Time and IoU with and without Fuzzy Mask as an Additional Input Prompt}
		\label{tab:time-comparison}
	\end{table}
	
	\subsection{Qualitative Evaluation}
	
	\label{subsec:qualitative-eval}
\begin{figure}[h!]
		\centering
		\label{fig:fuzzy-mask-samples}
		\begin{minipage}[c]{0.7\linewidth}
			\includegraphics[width=\linewidth]{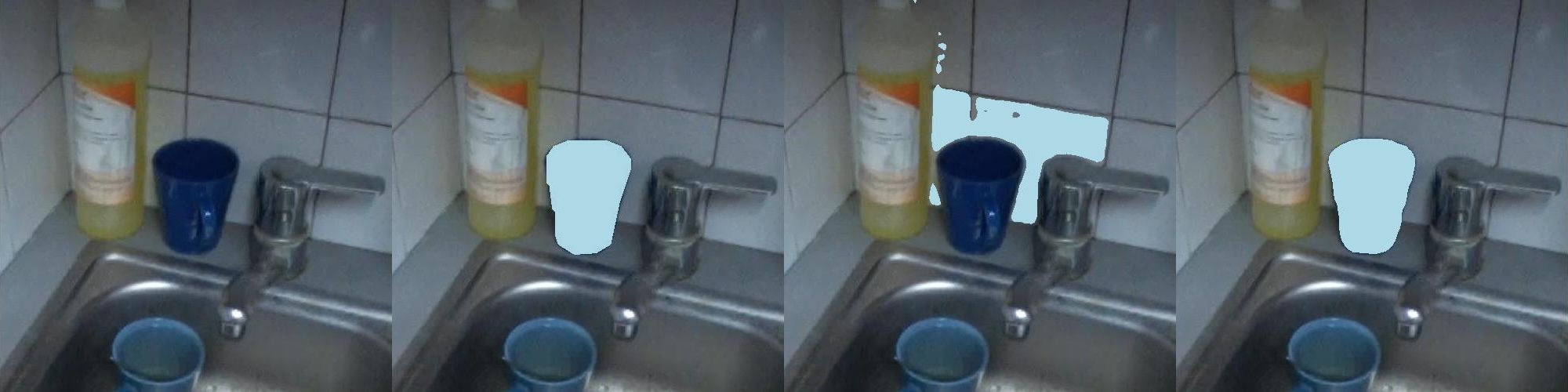}
			% \caption{}
		%	\label{fig:fuzzy-mask-sample}
		\end{minipage}
		% \hfill
		\begin{minipage}[c]{0.7\linewidth}
			\includegraphics[width=\linewidth]{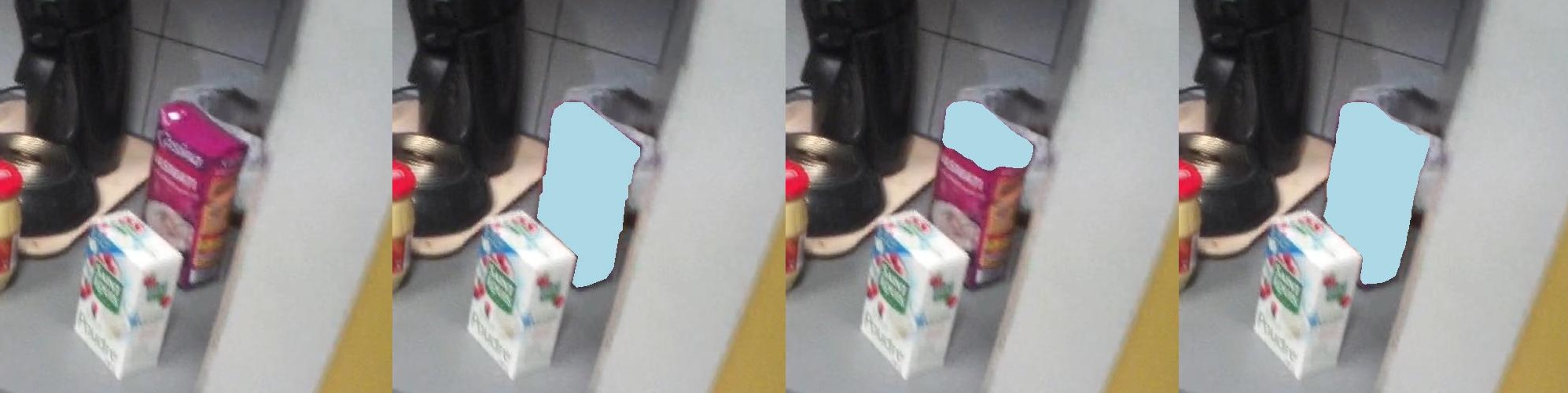}
			% \caption{}
	%		\label{fig:fuzzy-mask-sample}
		\end{minipage}
		\begin{minipage}[c]{0.7\linewidth}
			\includegraphics[width=\linewidth]{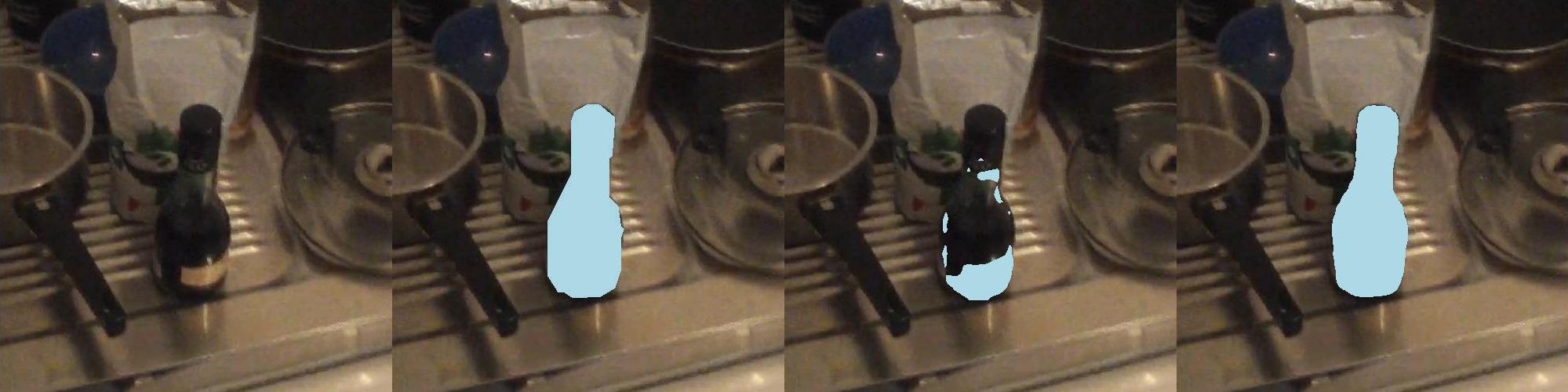}
			% \caption{}
		%	\label{fig:fuzzy-mask-sample}
		\end{minipage}
		% \hfill
		\begin{minipage}[c]{0.7\linewidth}
			\includegraphics[width=\linewidth]{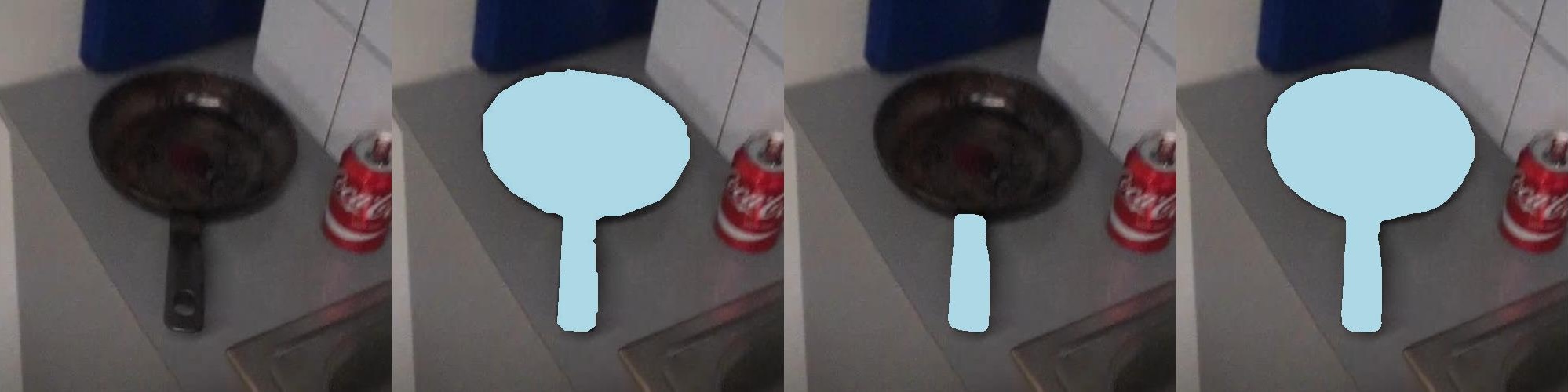}
			% \caption{}
	%		\label{fig:fuzzy-mask-sample}
		\end{minipage}
		\begin{minipage}[c]{0.7\linewidth}
			\includegraphics[width=\linewidth]{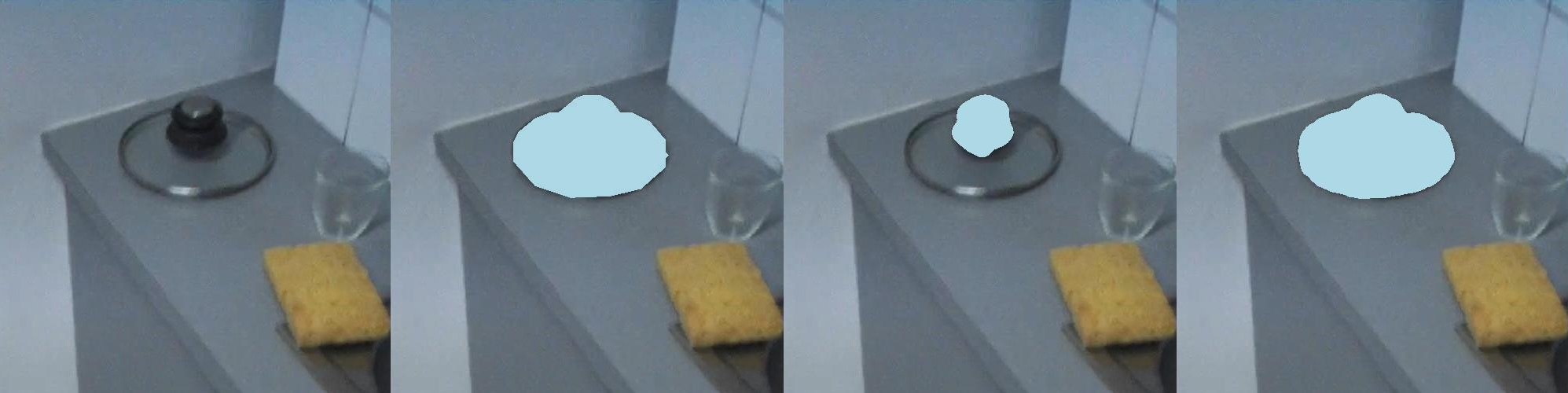}
			% \caption{}
		%	\label{fig:fuzzy-mask-sample}
		\end{minipage}
		% \hfill
		\begin{minipage}[c]{0.7\linewidth}
			\includegraphics[width=\linewidth]{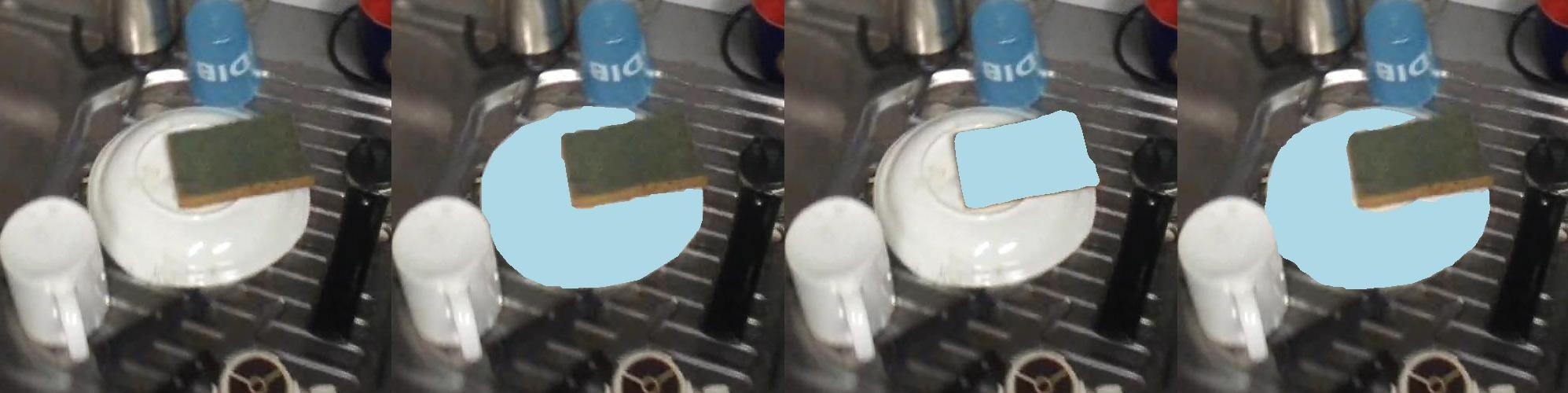}
			% \caption{}
			\label{fig:fuzzy-mask-sample}
		\end{minipage}
		
		\caption{Examples of segmented masks. From Left to Right: original frame, ground truth mask, mask from the pre-trained model, mask from fine-tuned model.}
		\label{fig:qualitative-eval}
	\end{figure}

	In figure \ref{fig:qualitative-eval}, we present some examples when SAM was fine-tuned on target class with binary masks only. 
	
	Compared to the ground truth masks depicted in the second column of figure  \ref{fig:qualitative-eval}, the pre-trained SAM model makes segmentation errors, see column 3. On the contrary, fine-tuned SAM gives totally consistent segmentation (column 4 of figure \ref{fig:qualitative-eval}).   \\
	
	\subsection{Real-World Scenario Testing}
	\label{subsec:real-world-testing}
	%we split by videos. 
	In this section, we aim to reproduce a real-world scenario and evaluate the proposed approaches under these conditions.
	Due to the sample size of the available dataset, different kitchen environments and subjects could not be separately considered for training and testing. Instead, different subjects were used for the same scenes, with the consideration that the variability in human physical characteristics, body part movement, overall movement and gaze fixations will allow the testing of real-world scenarios. 
	
	In contrast to previous experiments, where training and test data were randomly selected in a 70\% / 30\% split, whole video sequences are now used and selected per participant. Thus, whole video sequences and gaze fixations are selected for training for a subset of participants, and the system is evaluated on the complement of the selected subset. In our case, training is performed on participants 1,2,3 and evaluation is performed on participants 4 and 5. This assesses the robustness of the model and whether it is able to generalize across different individuals to avoid overfitting to specific video frames, especially if similar patterns are present in both the training and test sets.

	\subsubsection{Evaluation of fine-tuned models}
	\label{subsubsec:fine-tuned-models}
\begin{figure}[h!]
		\centering
		\includegraphics[width=0.9\linewidth]{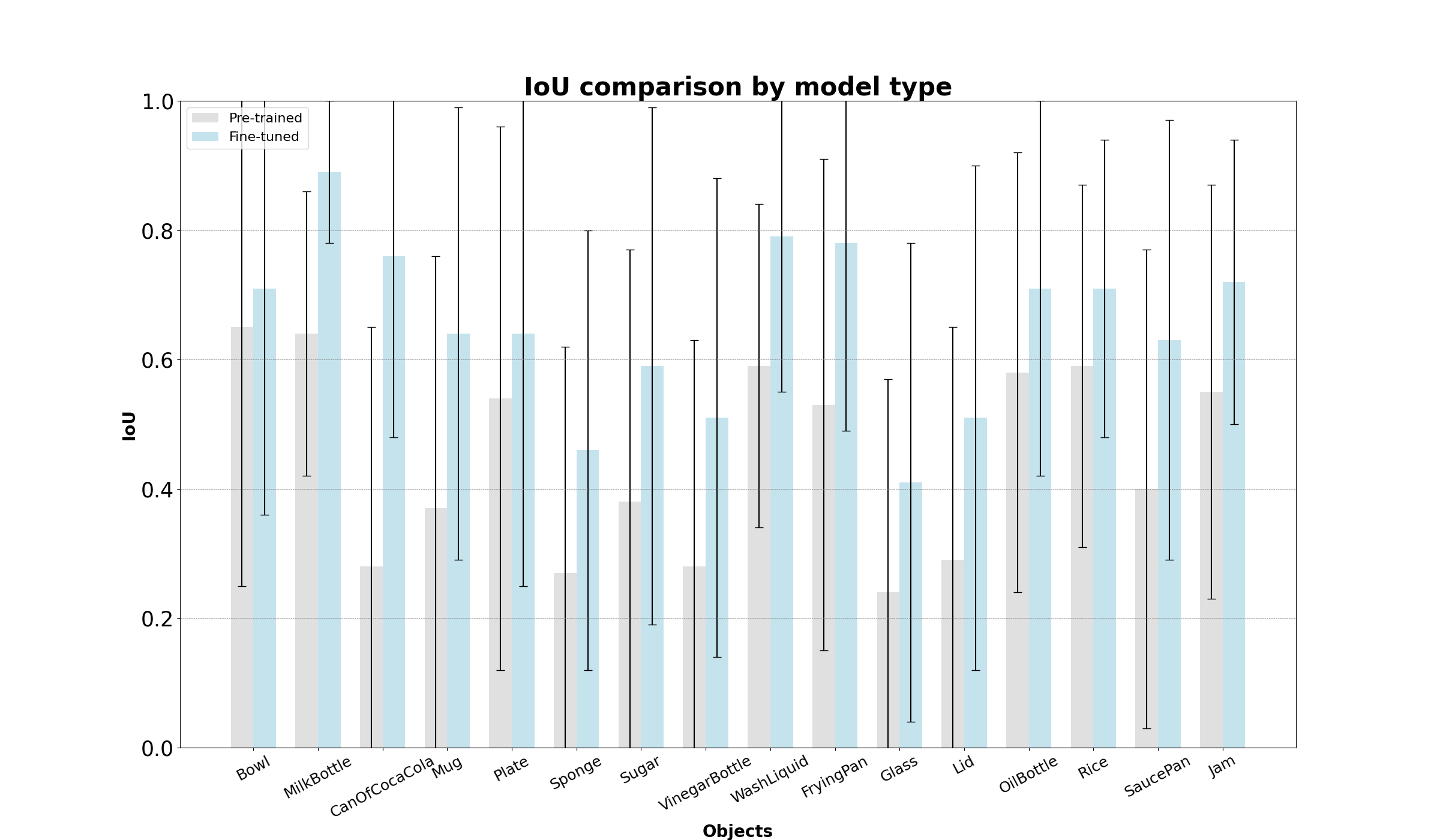}
		\caption{IoU evaluation comparison of pre-trained and fine-tuned models.}
		\label{fig:res-bin-mask-fuzzy-mask-video}
	\end{figure}

	Pre-trained and fine-tuned models are evaluated and compared in figure \ref{fig:res-bin-mask-fuzzy-mask-video}.
	
	Results illustrated in figure \ref{fig:res-bin-mask-fuzzy-mask-video} indicate that fine-tuning improves overall model performance, although less than that obtained in figure \ref{fig:res-pre-trained-bin-mask} using data from all participants. This reduced performance is due to the following factors. First, gaze fixations vary across individuals when viewing the same scene. Factors such as attention, experience, and cognitive processes influence where a person looks and cause variability in the training and test data, limiting the model's ability to generalize \cite{holmqvist2017eye, Kowler2011}. Second, people interact with objects and their environment differently, with hand movements and grasp angles that introduce variation and noise into the training data, limiting model generalization \cite{Fitts1992, newell1993variability}. Finally, human motion is inherently variable due to differences in speed, force, and trajectory for each individual \cite{bernshtein1967co, ROSENBAUM201093}. The variability between training and test subjects makes it difficult for the model to generalize well.
	
	\subsection{Ablation Study}
	\label{subsec:ablation}
\begin{figure}[h!]
		\centering
		\includegraphics[width=0.8\textwidth]{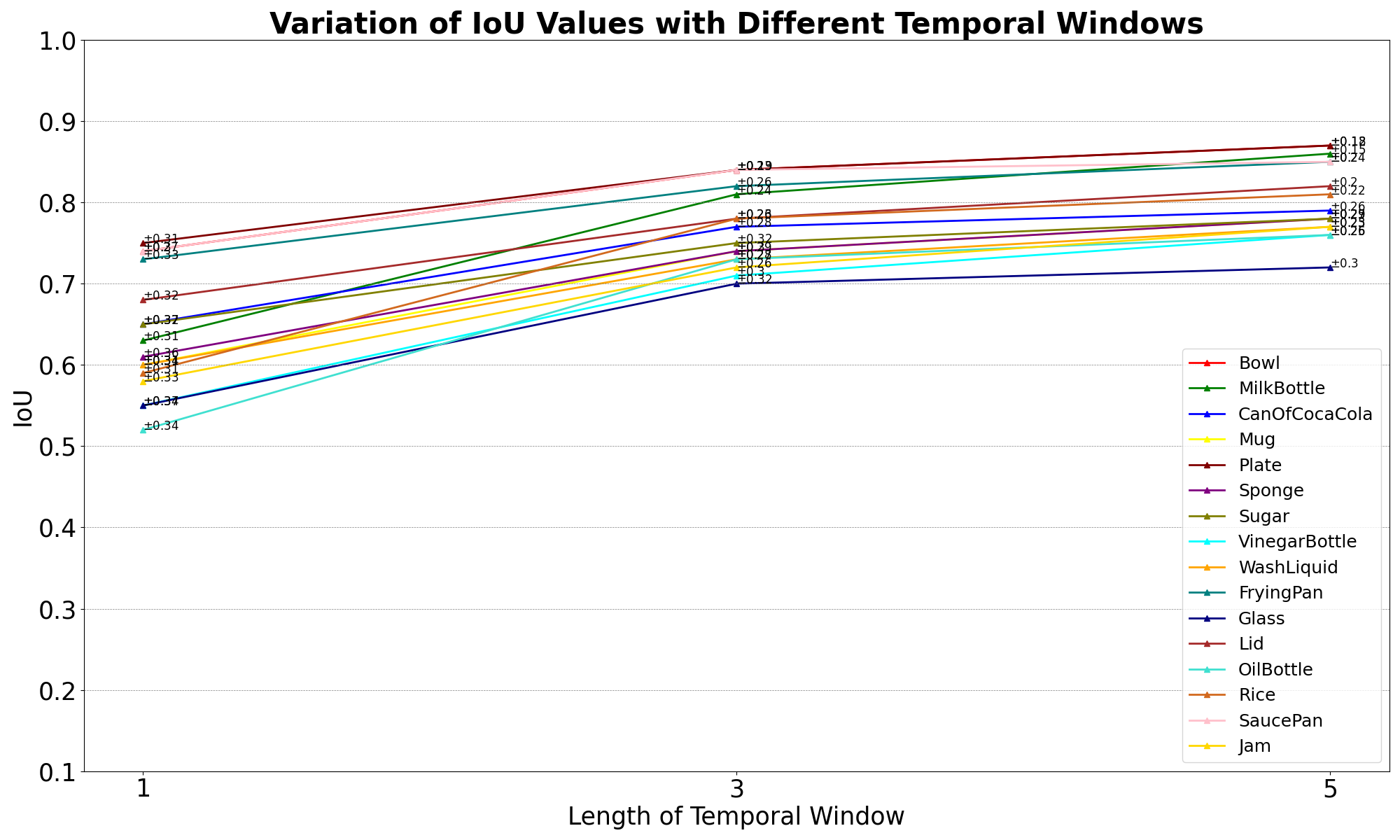}
		\caption{Influence of temporal window length on IoU value}
		\label{fig:ablation-study-temporal-window}
	\end{figure}

	The ablation study investigated the influence of the number of object points on the segmentation performance. It also investigated the possibility of eliminating the need for annotating the ground truth object masks by using fuzzy masks computed in section \ref{subsec:fuzzy-mask-computation} as ground truth masks for model fine-tuning. 
	The influence of object points was tested by varying the length of the temporal window between 1 and 5 frames. The results are shown in figure \ref{fig:ablation-study-temporal-window}. An increase in model performance is seen in figure \ref{fig:ablation-study-temporal-window} as the temporal window size increases, suggesting that prediction performance is a function of the number of object points used. Nevertheless, strongly increasing temporal window leads to the computational overload, due to the composition of homographies, see equation \ref{eqn:homography-chaining}, but also the need of estimation of pair-wise homography parameters for higher number of frames. 

\begin{figure}[h!]
		\centering
		\includegraphics[width=0.9\textwidth]{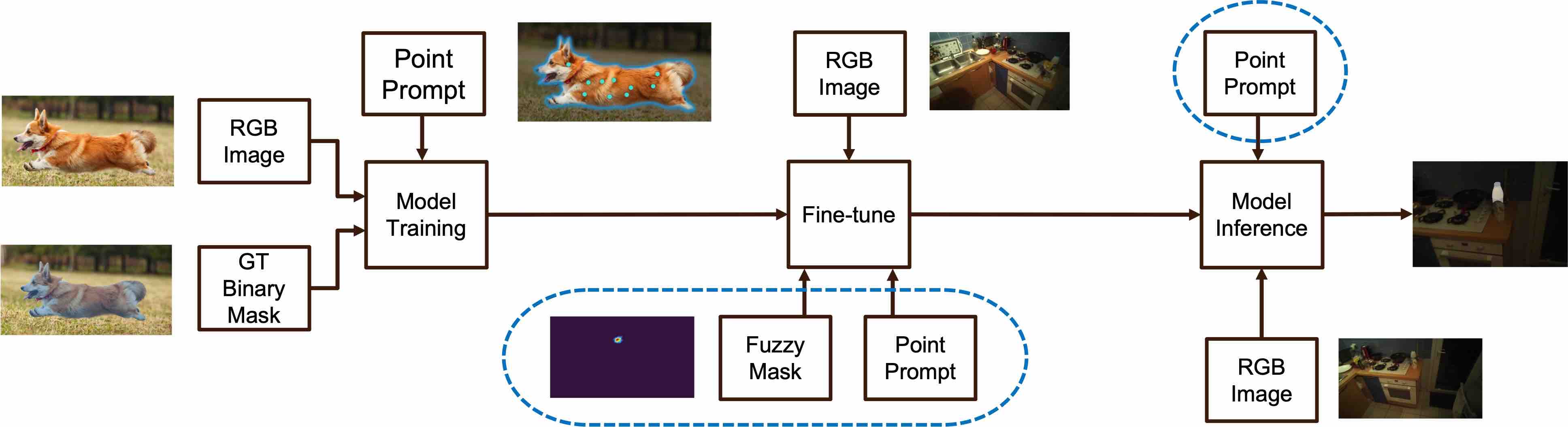}
		\caption{Application of SAM \cite{Kirillov_2023_ICCV} fine-tuned on GITW dataset}
		\label{fig:sam-application-fuzzy-mask-gt}
	\end{figure}

	The second ablation study considers a simplification of the segmentation process and proposes the use of automatically computed fuzzy masks as ground truth masks during model training. This eliminates the need to annotate binary ground-truth masks for each object added to the segmentation scenario, saving time and other resources. This application is illustrated in  figure \ref{fig:sam-application-fuzzy-mask-gt}.
	
	The evaluation of the proposal to substitute ground-truth masks is presented in terms of the mean intersection over union metric in figure \ref{fig:sam-application-fuzzy-mask-gt}. It can be observed that the segmentation performance is unsatisfactory across all object classes as the model instead learns to predict fuzzy masks, which do not represent the exact shape of the object.

	\begin{figure}[h!]
		\centering
		\includegraphics[width=0.8\textwidth]{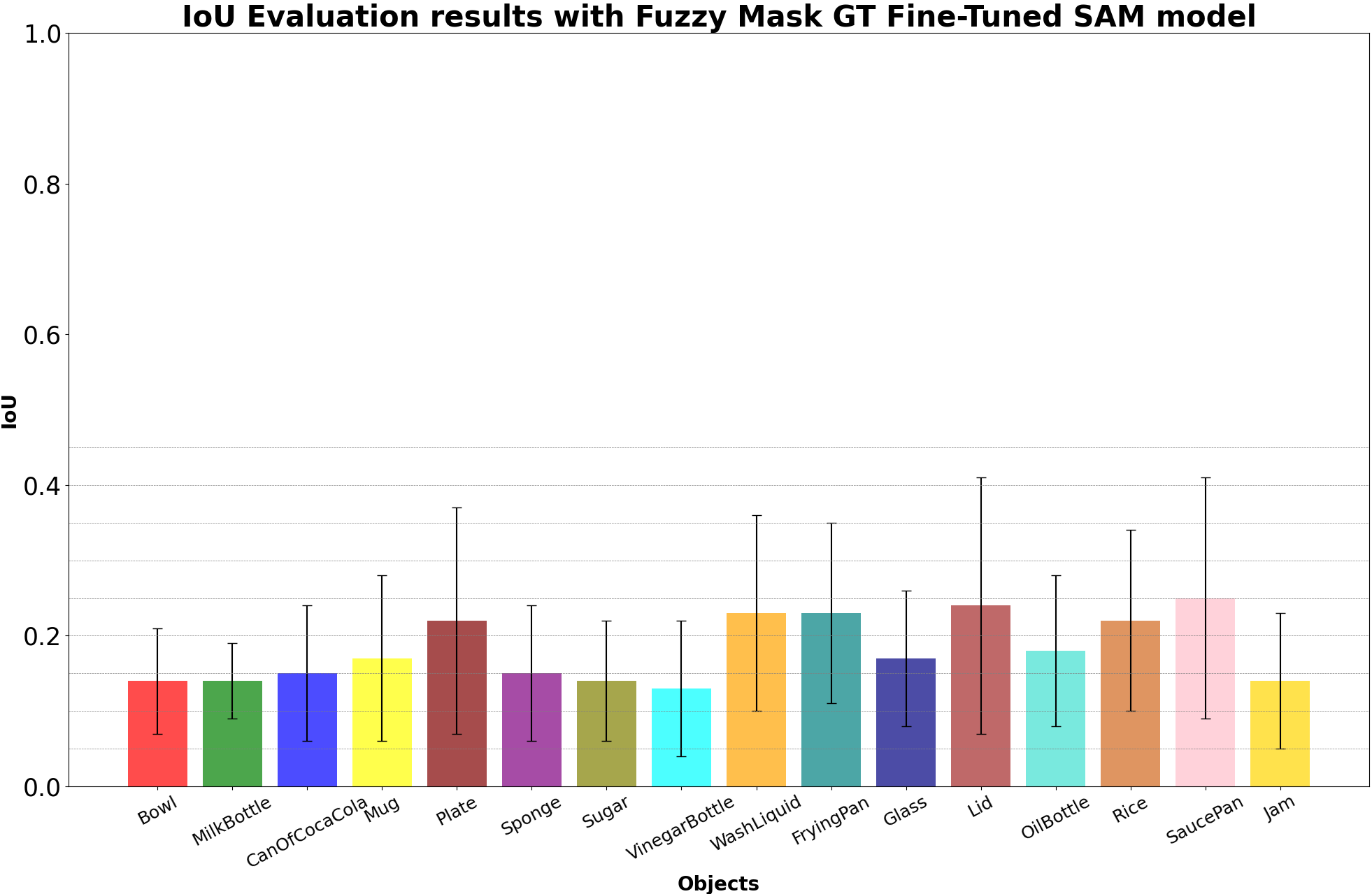}
		\caption{IoU evaluation result of model trained using fuzzy masks as ground-truth binary masks}
		\label{fig:res:ablation-study-fuzzy-mask}
	\end{figure}
	
	\section{Conclusion and perspectives}
	\label{sec:conclusion}
	In this paper, we have proposed an adaptation of the foundational model SAM for object segmentation in real-world cluttered visual scenes. We have proposed the generation of prompts for the model using the subject's gaze fixations, filtered to remove noise. An additional prompt using fuzzy masks based on gaze fixations was also introduced.
	
	Our results show that using gaze fixation points as prompts to the model through temporal window gaze point projection improves performance in both cases: the use of gaze fixations only and the use of the fuzzy masks together with fixation points. The ablation study suggests that segmentation results are better when the number of gaze fixation points on the occluded object of interest is higher.
	
	Adding a supplementary prompt, such as a fuzzy mask, for training and generalization by SAM models also slightly improves the results - up to 5\% of mean IoU increase in  a specific class.
	
	Overall, we have shown that despite the claim "Segment Anything", the foundation model needs to be domain-adapted, even if the target dataset is from the same optical imaging acquisition modality. Furthermore, the fine-tuning is recommended and improves the segmentation quality metric $mIoU$ by up to 0.51 points, with a better stability of the results for the whole scheme with domain-specific prompts and domain adaptation of the model.
	
	%Our results show that using gaze fixation points as additional inputs to the model through temporal window gaze point projection improves performance in both cases. The ablation study indicates that segmentation results are better when the number of gaze fixation points on the occluded object of interest is higher.
	
	%Adding a supplementary prompt, such as a fuzzy mask for training and generalization by SAM models also slightly improves the results - up to 5\% of mean IoU increase in  a specific class. Nevertheless, computation of this supplementary prompt for the SAM model induces stronger computational overload due to the process of computation of fuzzy masks and might not be compatible with real-time requirements of neuroprotheses control. 
	
	Our segmentation proposal has also been tested in a real-world scenario where the system has to be used by different subjects. Although the performance is lower, it is still acceptable, and the segmentation errors that could lead to incorrect 3D object pose estimation could be compensated by the residual body motion of the prosthesis wearers to reach the object to be grasped. 
	
	Finally, we have made available on the Roboflow Platform an annotated real-world first-person view object dataset with semantic segmentation masks. 
	
	The future work will consist of the integration of the proposed segmentation into the whole chain of segmentation and pose estimation of the objects to be grasped using RGB-D acquisition devices. 
	
	%	\section*{Acknowledgement}
	%The authors are acknowledging the  French National Agency of Research for its support.

	\section*{Availability of data and material}
	The annotated dataset\texttt{ Grasping In The Wild}, created and used in this paper, is freely available at: \url{https://universe.roboflow.com/iwrist/grasping-in-the-wild}
	\section*{Competing interests}
	Bolutife Atoki reports financial support was provided by French National Agency of Research. Other authors declare that they have no known competing financial interests or personalrelationships that could have appeared to influence the work reported in this paper.
	\section*{Funding}
	This research is supported by French National Agency of Research under grant agreement ANR I-WRIST-2024-2027
	\section*{Authors' contributions}

	 All authors have participated to the work presented in this manuscript. All authors have reviewed the manuscript and have approved the version to be submitted.

	% % To print the credit authorship contribution details
	
%\bibliographystyle{sn-mathphys-num}
%\bibliographystyle{cas-model2-names}

\bibliography{cas-refs}

\begin{thebibliography}{10}
\providecommand{\doi}[1]{\url{https://doi.org/#1}}
\bibcommenthead

\bibitem[\protect\citeauthoryear{Akcay and Breckon}{2017}]{8296499}
Akcay S, Breckon TP.
\newblock An evaluation of region based object detection strategies within
  X-ray baggage security imagery.
\newblock In: ICIP. IEEE; 2017. p. 1337--1341.

\bibitem[\protect\citeauthoryear{Chen et~al.}{2022}]{DBLP:conf/cvpr/ChenD022}
Chen Y, Dai H, Ding Y.
\newblock Pseudo-Stereo for Monocular 3D Object Detection in Autonomous
  Driving.
\newblock In: {CVPR}. {IEEE}; 2022. p. 877--887.

\bibitem[\protect\citeauthoryear{Xu et~al.}{2024}]{10323166}
Xu Z, Zhan X, Xiu Y, Suzuki C, Shimada K.
\newblock Onboard Dynamic-Object Detection and Tracking for Autonomous Robot
  Navigation With RGB-D Camera.
\newblock IEEE Robotics and Automation Letters. 2024;9(1):651--658.
\newblock \doi{10.1109/LRA.2023.3334683}.

\bibitem[\protect\citeauthoryear{Zhong
  et~al.}{2022}]{DBLP:journals/tcyb/ZhongHL22}
Zhong B, Huang H, Lobaton EJ.
\newblock Reliable Vision-Based Grasping Target Recognition for Upper Limb
  Prostheses.
\newblock {IEEE} Trans Cybern. 2022;52(3):1750--1762.
\newblock \doi{10.1109/TCYB.2020.2996960}.

\bibitem[\protect\citeauthoryear{Segas et~al.}{2023}]{segas_intuitive_2023}
Segas E, Mick S, Leconte V, Dubois O, Klotz R, Cattaert D, et~al.
\newblock Intuitive movement-based prosthesis control enables arm amputees to
  reach naturally in virtual reality.
\newblock eLife. 2023 Oct;12:RP87317.
\newblock \doi{10.7554/eLife.87317}.

\bibitem[\protect\citeauthoryear{González-Díaz
  et~al.}{2019}]{gonzalez-diaz_perceptually-guided_2019}
González-Díaz I, Benois-Pineau J, Domenger JP, Cattaert D, De~Rugy A.
\newblock Perceptually-guided deep neural networks for ego-action prediction:
  {Object} grasping.
\newblock Pattern Recognition. 2019 Apr;88:223--235.
\newblock \doi{10.1016/j.patcog.2018.11.013}.

\bibitem[\protect\citeauthoryear{Wang et~al.}{2019}]{wang_densefusion_2019}
Wang C, Xu D, Zhu Y, Martín-Martín R, Lu C, Fei-Fei L, et~al.: {DenseFusion}:
  {6D} {Object} {Pose} {Estimation} by {Iterative} {Dense} {Fusion}.
\newblock arXiv.
\newblock Version Number: 1.
\newblock Available from: \url{https://arxiv.org/abs/1901.04780}.

\bibitem[\protect\citeauthoryear{Chen
  et~al.}{2018}]{DBLP:journals/pami/ChenPKMY18}
Chen L, Papandreou G, Kokkinos I, Murphy K, Yuille AL.
\newblock DeepLab: Semantic Image Segmentation with Deep Convolutional Nets,
  Atrous Convolution, and Fully Connected CRFs.
\newblock {IEEE} Trans Pattern Anal Mach Intell. 2018;40(4):834--848.

\bibitem[\protect\citeauthoryear{Bommasani
  et~al.}{2021}]{DBLP:journals/corr/abs-2108-07258}
Bommasani R, Hudson DA, Adeli E, et~al RBA.
\newblock On the Opportunities and Risks of Foundation Models.
\newblock CoRR. 2021;abs/2108.07258.
\newblock {\href{https://arxiv.org/abs/2108.07258}{{2108.07258}}}.

\bibitem[\protect\citeauthoryear{Wu et~al.}{2019}]{wu2019detectron2}
Wu Y, Kirillov A, Massa F, Lo WY, Girshick R.: Detectron2.
\newblock \url{https://github.com/facebookresearch/detectron2}.

\bibitem[\protect\citeauthoryear{Kirillov et~al.}{2023}]{Kirillov_2023_ICCV}
Kirillov A, Mintun E, Ravi N, Mao H, Rolland C, Gustafson L, et~al.
\newblock Segment Anything.
\newblock In: {ICCV}. {IEEE}; 2023. p. 3992--4003.

\bibitem[\protect\citeauthoryear{Chassery and Garbay}{1984}]{ChasseryG84}
Chassery J, Garbay C.
\newblock An Iterative Segmentation Method Based on a Contextual Color and
  Shape Criterion.
\newblock {IEEE} Trans Pattern Anal Mach Intell. 1984;6(6):794--800.

\bibitem[\protect\citeauthoryear{Ciresan
  et~al.}{2012}]{DBLP:conf/nips/CiresanGGS12}
Ciresan DC, Giusti A, Gambardella LM, Schmidhuber J.
\newblock Deep Neural Networks Segment Neuronal Membranes in Electron
  Microscopy Images.
\newblock In: {NIPS}; 2012. p. 2852--2860.

\bibitem[\protect\citeauthoryear{Krizhevsky
  et~al.}{2012}]{DBLP:conf/nips/KrizhevskySH12}
Krizhevsky A, Sutskever I, Hinton GE.
\newblock ImageNet Classification with Deep Convolutional Neural Networks.
\newblock In: {NIPS}; 2012. p. 1106--1114.

\bibitem[\protect\citeauthoryear{Simonyan and
  Zisserman}{2015}]{DBLP:journals/corr/SimonyanZ14a}
Simonyan K, Zisserman A.
\newblock Very Deep Convolutional Networks for Large-Scale Image Recognition.
\newblock In: {ICLR}; 2015. .

\bibitem[\protect\citeauthoryear{Szegedy
  et~al.}{2015}]{DBLP:conf/cvpr/SzegedyLJSRAEVR15}
Szegedy C, Liu W, Jia Y, Sermanet P, Reed SE, Anguelov D, et~al.
\newblock Going deeper with convolutions.
\newblock In: {CVPR}. {IEEE} Computer Society; 2015. p. 1--9.

\bibitem[\protect\citeauthoryear{Ronneberger}{2017}]{DBLP:conf/bildmed/Ronneberger17}
Ronneberger O.
\newblock Invited Talk: U-Net Convolutional Networks for Biomedical Image
  Segmentation.
\newblock In: Bildverarbeitung f{\"{u}}r die Medizin. Informatik Aktuell.
  Springer; 2017. p.~3.

\bibitem[\protect\citeauthoryear{Badrinarayanan
  et~al.}{2017}]{DBLP:journals/pami/BadrinarayananK17}
Badrinarayanan V, Kendall A, Cipolla R.
\newblock SegNet: {A} Deep Convolutional Encoder-Decoder Architecture for Image
  Segmentation.
\newblock {IEEE} Trans Pattern Anal Mach Intell. 2017;39(12):2481--2495.

\bibitem[\protect\citeauthoryear{Chen et~al.}{2018}]{DBLP:conf/eccv/ChenZPSA18}
Chen L, Zhu Y, Papandreou G, Schroff F, Adam H.
\newblock Encoder-Decoder with Atrous Separable Convolution for Semantic Image
  Segmentation.
\newblock In: {ECCV} {(7)}. vol. 11211 of Lecture Notes in Computer Science.
  Springer; 2018. p. 833--851.

\bibitem[\protect\citeauthoryear{Deng
  et~al.}{2009}]{DBLP:conf/cvpr/DengDSLL009}
Deng J, Dong W, Socher R, Li L, Li K, Fei{-}Fei L.
\newblock ImageNet: {A} large-scale hierarchical image database.
\newblock In: {CVPR}. {IEEE} Computer Society; 2009. p. 248--255.

\bibitem[\protect\citeauthoryear{Lin
  et~al.}{2014}]{DBLP:conf/eccv/LinMBHPRDZ14}
Lin T, Maire M, Belongie SJ, Hays J, Perona P, Ramanan D, et~al.
\newblock Microsoft {COCO:} Common Objects in Context.
\newblock In: {ECCV} {(5)}. vol. 8693 of Lecture Notes in Computer Science.
  Springer; 2014. p. 740--755.

\bibitem[\protect\citeauthoryear{Jocher}{2020}]{yolov5}
Jocher G.: Ultralytics YOLOv5.
\newblock Available from: \url{https://github.com/ultralytics/yolov5}.

\bibitem[\protect\citeauthoryear{Venkatesh
  et~al.}{2024}]{DBLP:journals/sncs/VenkateshAGYP24}
Venkatesh R, Anantharajan S, Gunasekaran S, Yogaraja CA, Poornima IGA.
\newblock Prediction of Alzheimer's Disease Using Adaptive Fine-Tuned Deep
  Resnet-50 with Attention Mechanism.
\newblock {SN} Comput Sci. 2024;5(4):392.

\bibitem[\protect\citeauthoryear{Liu et~al.}{2023}]{DBLP:conf/mm/LiuZGZ23}
Liu H, Zhang L, Guan J, Zhou S.
\newblock Zero-Shot Object Detection by Semantics-Aware {DETR} with Adaptive
  Contrastive Loss.
\newblock In: {ACM} Multimedia. {ACM}; 2023. p. 4421--4430.

\bibitem[\protect\citeauthoryear{Bommasani
  et~al.}{2021}]{bommasani2021opportunities}
Bommasani R, Hudson DA, Adeli E, Altman R, Arora S, von Arx S, et~al.
\newblock On the opportunities and risks of foundation models.
\newblock arXiv preprint arXiv:210807258. 2021;.

\bibitem[\protect\citeauthoryear{Awais
  et~al.}{2023}]{DBLP:journals/corr/abs-2307-13721}
Awais M, Naseer M, Khan SH, Anwer RM, Cholakkal H, Shah M, et~al.
\newblock Foundational Models Defining a New Era in Vision: {A} Survey and
  Outlook.
\newblock CoRR. 2023;abs/2307.13721.

\bibitem[\protect\citeauthoryear{Zhang et~al.}{2024}]{DBLP:conf/cvpr/ZhangDS24}
Zhang Y, Doughty H, Snoek CGM.
\newblock Low-Resource Vision Challenges for Foundation Models.
\newblock In: {CVPR}. {IEEE}; 2024. p. 21956--21966.

\bibitem[\protect\citeauthoryear{Oquab
  et~al.}{2024}]{DBLP:journals/tmlr/OquabDMVSKFHMEA24}
Oquab M, Darcet T, Moutakanni T, Vo HV, Szafraniec M, Khalidov V, et~al.
\newblock DINOv2: Learning Robust Visual Features without Supervision.
\newblock Trans Mach Learn Res. 2024;2024.

\bibitem[\protect\citeauthoryear{Cheng
  et~al.}{2022}]{DBLP:conf/cvpr/ChengMSKG22}
Cheng B, Misra I, Schwing AG, Kirillov A, Girdhar R.
\newblock Masked-attention Mask Transformer for Universal Image Segmentation.
\newblock In: {CVPR}. {IEEE}; 2022. p. 1280--1289.

\bibitem[\protect\citeauthoryear{Cheng et~al.}{2021}]{DBLP:conf/nips/ChengSK21}
Cheng B, Schwing AG, Kirillov A.
\newblock Per-Pixel Classification is Not All You Need for Semantic
  Segmentation.
\newblock In: NeurIPS; 2021. p. 17864--17875.

\bibitem[\protect\citeauthoryear{He et~al.}{2016}]{DBLP:conf/cvpr/HeZRS16}
He K, Zhang X, Ren S, Sun J.
\newblock Deep Residual Learning for Image Recognition.
\newblock In: {CVPR}. {IEEE} Computer Society; 2016. p. 770--778.

\bibitem[\protect\citeauthoryear{Liu
  et~al.}{2021}]{DBLP:conf/iccv/LiuL00W0LG21}
Liu Z, Lin Y, Cao Y, Hu H, Wei Y, Zhang Z, et~al.
\newblock Swin Transformer: Hierarchical Vision Transformer using Shifted
  Windows.
\newblock In: {ICCV}. {IEEE}; 2021. p. 9992--10002.

\bibitem[\protect\citeauthoryear{Xie et~al.}{2017}]{DBLP:conf/cvpr/XieGDTH17}
Xie S, Girshick RB, Doll{\'{a}}r P, Tu Z, He K.
\newblock Aggregated Residual Transformations for Deep Neural Networks.
\newblock In: {CVPR}. {IEEE} Computer Society; 2017. p. 5987--5995.

\bibitem[\protect\citeauthoryear{Obeso et~al.}{2022}]{ObesoBGR22}
Obeso AM, Benois{-}Pineau J, Garc{\'{\i}}a{-}V{\'{a}}zquez MS,
  Ram{\'{\i}}rez{-}Acosta AA.
\newblock Visual vs internal attention mechanisms in deep neural networks for
  image classification and object detection.
\newblock Pattern Recognit. 2022;123:108411.

\bibitem[\protect\citeauthoryear{Lento et~al.}{2024}]{Lento2024.01.30.577386}
Lento B, Segas E, Leconte V, Doat E, Danion F, P{\'e}teri R, et~al.
\newblock 3D-ARM-Gaze: a public dataset of 3D Arm Reaching Movements with Gaze
  information in virtual reality.
\newblock bioRxiv. 2024;\doi{10.1101/2024.01.30.577386}.

\bibitem[\protect\citeauthoryear{Dosovitskiy
  et~al.}{2021}]{DBLP:conf/iclr/DosovitskiyB0WZ21}
Dosovitskiy A, Beyer L, Kolesnikov A, Weissenborn D, Zhai X, Unterthiner T,
  et~al.
\newblock An Image is Worth 16x16 Words: Transformers for Image Recognition at
  Scale.
\newblock In: {ICLR}. OpenReview.net; 2021. .

\bibitem[\protect\citeauthoryear{Vaswani
  et~al.}{2017}]{DBLP:conf/nips/VaswaniSPUJGKP17}
Vaswani A, Shazeer N, Parmar N, Uszkoreit J, Jones L, Gomez AN, et~al.
\newblock Attention is All you Need.
\newblock In: {NIPS}; 2017. p. 5998--6008.

\bibitem[\protect\citeauthoryear{Unser}{1999}]{799930}
Unser M.
\newblock Splines: a perfect fit for signal and image processing.
\newblock IEEE Signal Processing Magazine. 1999;16(6):22--38.
\newblock \doi{10.1109/79.799930}.

\bibitem[\protect\citeauthoryear{Kadner
  et~al.}{2023}]{DBLP:conf/wacv/KadnerTHR23}
Kadner F, Thomas T, Hoppe D, Rothkopf CA.
\newblock Improving saliency models' predictions of the next fixation with
  humans' intrinsic cost of gaze shifts.
\newblock In: {WACV}. {IEEE}; 2023. p. 2103--2113.

\bibitem[\protect\citeauthoryear{Fej{\'{e}}r
  et~al.}{2022}]{DBLP:journals/jimaging/FejerNBSRD22}
Fej{\'{e}}r A, Nagy Z, Benois{-}Pineau J, Szolgay P, de~Rugy A, Domenger J.
\newblock Hybrid FPGA-CPU-Based Architecture for Object Recognition in Visual
  Servoing of Arm Prosthesis.
\newblock J Imaging. 2022;8(2):44.

\bibitem[\protect\citeauthoryear{Silverman}{2018}]{silverman2018density}
Silverman BW.
\newblock Density estimation for statistics and data analysis.
\newblock Routledge; 2018.

\bibitem[\protect\citeauthoryear{Otero-Millan et~al.}{2008}]{10.1167/8.14.21}
Otero-Millan J, Troncoso XG, Macknik SL, Serrano-Pedraza I, Martinez-Conde S.
\newblock {Saccades and microsaccades during visual fixation, exploration, and
  search: Foundations for a common saccadic generator}.
\newblock Journal of Vision. 2008 12;8(14):21--21.
\newblock \doi{10.1167/8.14.21}.

\bibitem[\protect\citeauthoryear{{LaBRI}}{}]{graspingInTheWild}
{LaBRI}.: Grasping-in-the-Wild Dataset.
\newblock Accessed: 2024-07-11.
\newblock \url{https://www.labri.fr/projet/AIV/graspinginthewild.php}.

\bibitem[\protect\citeauthoryear{{Tobii Technology}}{}]{tobiiProGlasses2}
{Tobii Technology}.: Tobii Pro Glasses 2.
\newblock Accessed: 2024-07-11.
\newblock
  \url{https://www.tobii.com/products/discontinued/tobii-pro-glasses-2}.

\bibitem[\protect\citeauthoryear{M{\"u}ller et~al.}{2022}]{Muller2022}
M{\"u}ller D, Soto-Rey I, Kramer F.
\newblock Towards a guideline for evaluation metrics in medical image
  segmentation.
\newblock BMC Research Notes. 2022 Jun;15(1):210.
\newblock \doi{10.1186/s13104-022-06096-y}.

\bibitem[\protect\citeauthoryear{Holmqvist and
  Andersson}{2017}]{holmqvist2017eye}
Holmqvist K, Andersson R.
\newblock Eye Tracking: A Comprehensive Guide to Methods, Paradigms, and
  Measures.
\newblock Lund Eye-Tracking Research Institute; 2017.

\bibitem[\protect\citeauthoryear{Kowler}{2011}]{Kowler2011}
Kowler E.
\newblock Eye movements: the past 25 years.
\newblock Vision Research. 2011 Jul;51(13):1457--1483.
\newblock \doi{10.1016/j.visres.2010.12.014}.
\newblock {\href{https://arxiv.org/abs/PMC3094591}{{PMC3094591}}}.

\bibitem[\protect\citeauthoryear{Fitts}{1992}]{Fitts1992}
Fitts PM.
\newblock The information capacity of the human motor system in controlling the
  amplitude of movement. 1954.
\newblock Journal of Experimental Psychology: General. 1992
  Sep;121(3):262--269.
\newblock Research Support, U.S. Gov't, Non-P.H.S..
  \doi{10.1037//0096-3445.121.3.262}.

\bibitem[\protect\citeauthoryear{Newell and
  Corcos}{1993}]{newell1993variability}
Newell KM, Corcos DM.
\newblock Variability and Motor Control.
\newblock Human Kinetics Publishers; 1993.

\bibitem[\protect\citeauthoryear{Bernshte{\u\i}n}{1967}]{bernshtein1967co}
Bernshte{\u\i}n NA.
\newblock The Co-ordination and Regulation of Movements.
\newblock Pergamon Press; 1967.

\bibitem[\protect\citeauthoryear{Rosenbaum}{2010}]{ROSENBAUM201093}
Rosenbaum DA.
\newblock Chapter 4 - Psychological Foundations.
\newblock In: Rosenbaum DA, editor. Human Motor Control (Second Edition).
  second edition ed. San Diego: Academic Press; 2010. p. 93--134.

\end{thebibliography}

\end{document}